\documentclass[10pt,twocolumn,letterpaper]{article}

\usepackage[pagenumbers]{cvpr} 

\definecolor{cvprblue}{rgb}{0.21,0.49,0.74}
\usepackage[pagebackref,breaklinks,colorlinks,allcolors=cvprblue]{hyperref}
\usepackage{lipsum}
\usepackage{multirow}
\usepackage{pifont}

\usepackage[table]{xcolor}
\usepackage{booktabs}     
\usepackage{tabularx}     
\usepackage{array}        
\usepackage{makecell}     
\usepackage{arydshln}
\usepackage{graphicx}
\usepackage{colortbl}
\usepackage{xcolor}
\usepackage{xparse}

\NewDocumentCommand{\cellimg}{O{\linewidth}m}{%
  \includegraphics[width=#1]{#2}%
}

\def\confName{CVPR}
\def\confYear{2026}

\title{Reasoning Guided Embeddings: Leveraging MLLM Reasoning \\ for Improved Multimodal Retrieval}

\author{%
  \hspace{-8mm} Chunxu Liu$^{1,2}$\thanks{Work is done during internship at Sensetime Research.} \quad Jiyuan Yang$^{2, 3}$ \quad Ruopeng Gao$^1$ \quad Yuhan Zhu$^1$ \quad Feng Zhu$^2$ \quad Rui Zhao$^2$\quad Limin Wang$^{1,4}\thanks{Corresponding author  (lmwang@nju.edu.cn).}$ \\
  $^1$State Key Laboratory for Novel Software Technology, Nanjing University\\
  $^2$Sensetime Research \quad $^3$ Beijing Institute of Technology \quad $^4$Shanghai AI Lab\\
  \newline
  \\
  \hspace{-10mm} \textbf{\url{https://github.com/MCG-NJU/RGE}}
}

\begin{document}
\maketitle
\begin{abstract}

Multimodal embeddings are widely used in downstream tasks such as multimodal retrieval, enabling alignment of interleaved modalities in a shared representation space. While recent studies show that Multimodal Large Language Models (MLLMs) can serve as strong embedding extractors, existing approaches treat embedding extraction as a direct encoding step, overlooking the fact that MLLMs possess the generative capability for reasoning that could be leveraged to enhance representation quality. In this work, we explore how to explicitly incorporate reasoning into the embedding process. To this end, we propose \textbf{R}easoning \textbf{G}uided \textbf{E}mbeddings (\textbf{RGE}), which preserves the generative rationale process of MLLMs and couples it with contrastive training. Our method first enables the model to perform structured rationale generation conditioned on the instruction, and then extracts representations after reasoning has unfolded. This simple design enhances the context-conditional inference signals within the embedding, leading to improved multimodal representation quality. Experiments on the MMEB benchmark show that reasoning-guided conditioning improves multimodal retrieval performance by 4.9\% over the non-reasoning baseline, confirming that explicit reasoning can effectively enhance embedding quality.
\end{abstract}    
\section{Introduction}
\label{sec:intro}

Multimodal embeddings are fundamental to modern vision–language systems, as they compress rich multimodal signals into dense vectors that can directly support a wide range of downstream tasks. Recently, extracting multimodal features from Multimodal Large Language Models (MLLMs) is emerging as a promising paradigm. Compared to CLIP-like two-tower dual encoders~\cite{radford2021learning, li2022blip, zhai2023sigmoid}, MLLMs offer native multi-modal fusion within a single generative architecture. Inspired by the success of using LLM as embedding generators~\cite{behnamghader2024llmvec, lee2025nvembed}, a growing line of work has extended this paradigm to multimodal large language models (MLLMs)~\cite{jiang2024e5, jiang2025vlmvec, liu2025lamra}, demonstrating that generative architectures can surpass contrastively trained encoders on retrieval and general-purpose embedding benchmarks.
\begin{figure}[t]
  \centering
  \includegraphics[width=\linewidth]{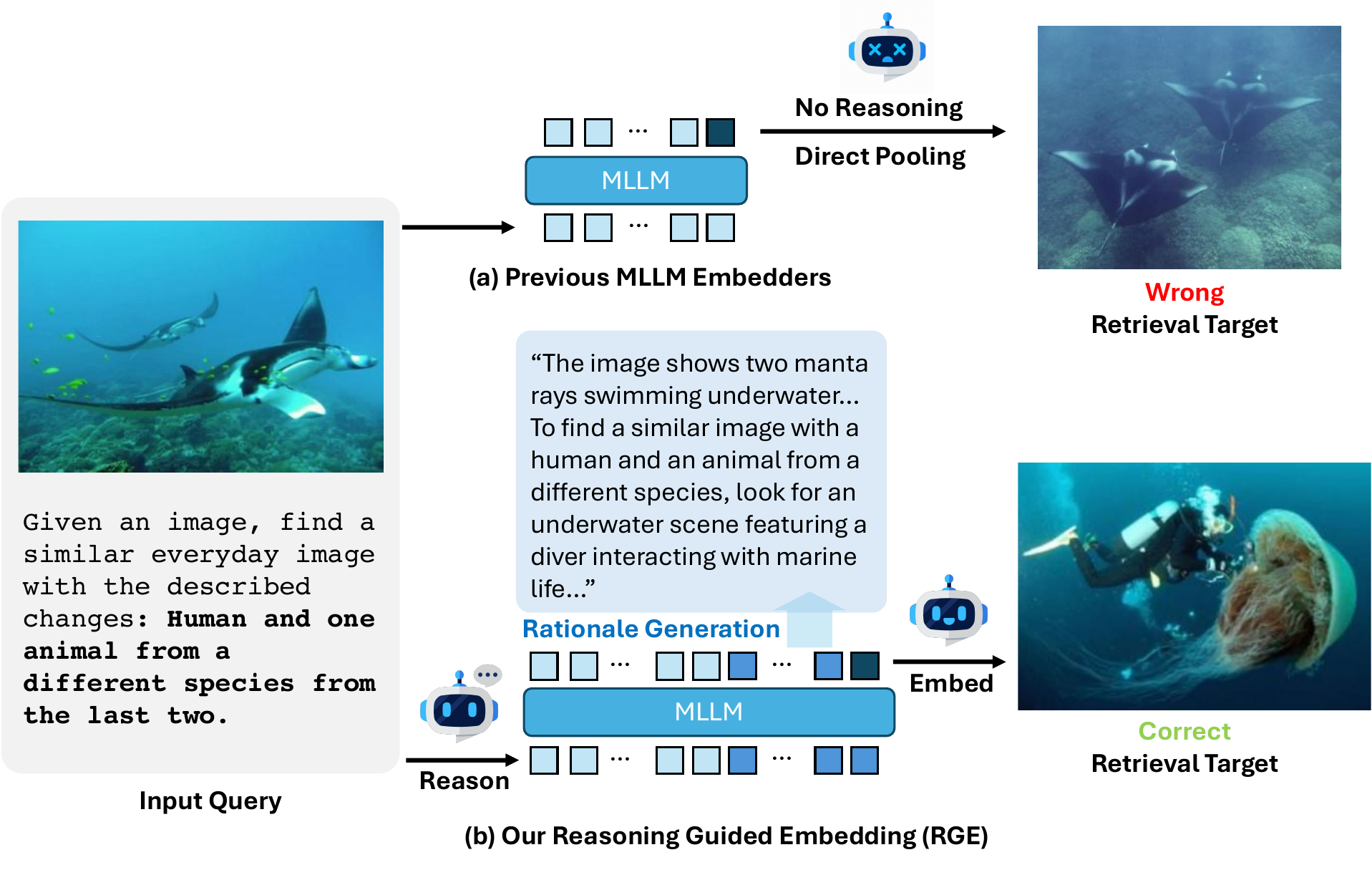}
  \caption{\textbf{Comparison between our method and non-reasoning baseline.} Unlike previous MLLM embedders that directly encode inputs into embeddings without reasoning, our MLLM embedding model (RGE) first performs reasoning over the input before embedding extraction, then uses both the query and its generated rationale to produce a more informative embedding. The illustrated example is a composed image retrieval task (text–image to image) from the CIRR~\cite{CIRR} subset in MMEB~\cite{jiang2025vlmvec} benchmark.}
  \label{fig:introfig}
  \vspace{-0.5cm}
\end{figure}

However, prevailing MLLM embedding pipelines typically take the last-layer hidden state of the final token as the semantic representation. This practice overlooks one of the key strengths of MLLMs, namely their ability for context-conditioned reasoning, which is reflected in intermediate representations shaped by multimodal inputs and instructions. Furthermore, many downstream tasks naturally require such reasoning capabilities, including fine-grained classification, visual question answering (VQA), instruction-grounded retrieval, visual grounding, and other settings where relevant evidence must be selected, organized, and justified. While MLLMs are capable to produce more accurate answers when explicitly prompted to reason~\cite{wei2022chain} or finetuned to reason~\cite{guo2025deepseek}, current MLLM embedding models ignore that capability: reasoning is neither invoked nor captured, and embeddings are extracted without regard to when and how the model reasons. 

Therefore, we propose to explicitly recover context-conditional reasoning for MLLM embeddings by preserving the model’s generative reasoning process. Instead of directly pooling hidden states from raw inputs, we enable the MLLM first to generate structured rationale and then extract representations within the same decoding context, after the model has formed an internal explanation of the input under the given instruction. This makes the embedding aware of the intended inference pattern. We argue that in scenarios where the instructions are compositional, domain-specific, or cognitively demanding, reasoning-guided conditioning becomes essential for extracting task-relevant features, serving as a form of test-time computation that allocates additional inference to refine the embedding quality.

Concretely, we design a joint training framework that couples rationale generation with contrastive supervision. Directly using hidden states from next-token prediction for contrastive alignment leads to information leakage, as the model can rely on oracle rationale tokens instead of genuine reasoning. To mitigate this, we supervise rationale generation with oracle rationales while performing contrastive learning on embeddings derived from self-generated rationales. This decoupled yet connected design links reasoning quality to representation quality, allowing the model to learn embeddings that are both more discriminative and more contextually grounded. As illustrated in \Cref{fig:introfig}, a baseline model trained without reasoning may over-focus on the surface form of the query and retrieve a misleading candidate.  In contrast, our model performs a reasoning step before embedding extraction, leveraging its world knowledge to capture deeper semantics and retrieve the correct candidate. Extensive experiments show that this rationale-first pipeline consistently improves representation quality across diverse datasets.

In conclusion, our main contributions are threefold:
\begin{enumerate}
    \item We propose Reasoning Guided Embedding (RGE), a simple design that makes MLLMs explicitly generate rationales before embedding extraction, resulting in more informative representations for retrieval.
    \item We propose a joint training framework that requires no architectural modification, combining language modeling and contrastive objectives. We identify an information leakage issue in naive contrastive supervision and address it by applying contrastive learning on self-generated rationales produced on-the-fly. We will release the oracle rationales, models, and code to support future research.
    \item On MMEB, RGE achieves state-of-the-art multimodal retrieval performance at comparable model scales, surpassing the non-reasoning baseline by 4.9\%, confirming its effectiveness.
\end{enumerate}

\section{Related Work}
\label{sec:related_work}

\begin{figure*}[t]
    \centering
    \includegraphics[width=\linewidth]{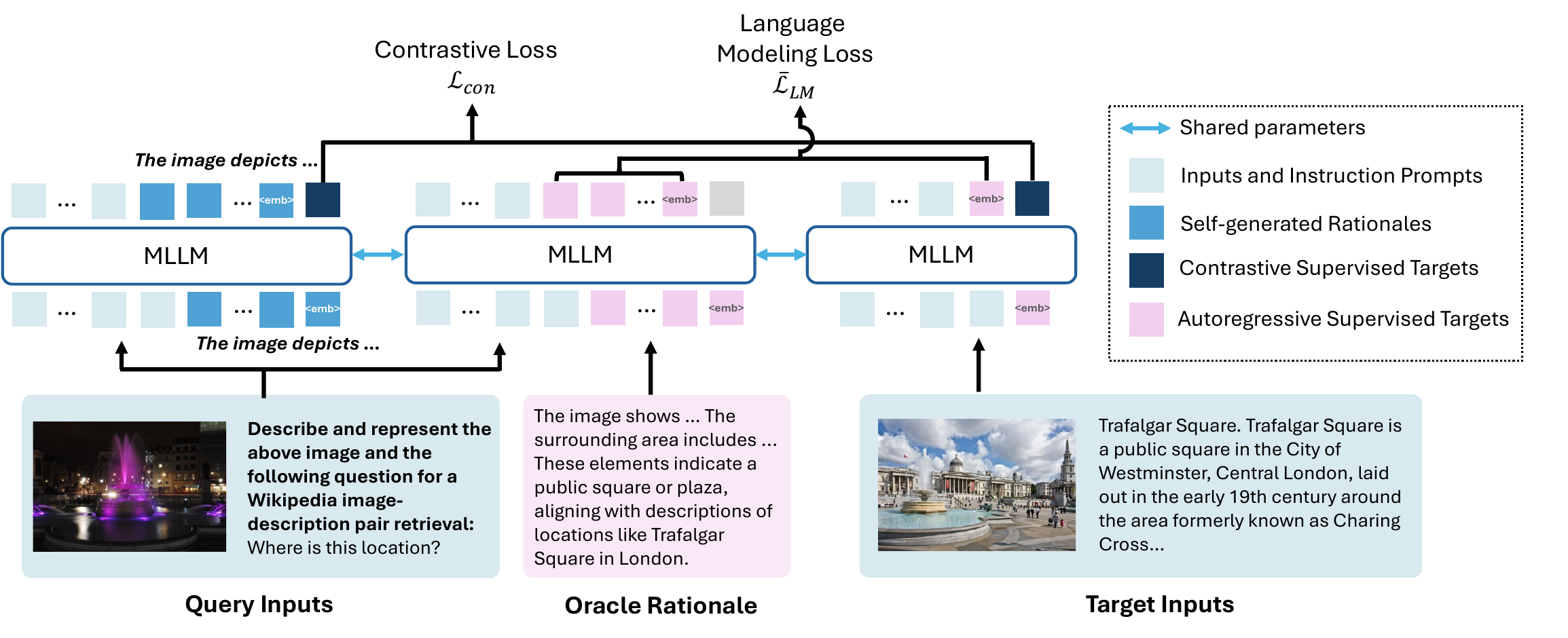}
    \caption{\textbf{Overview of our training pipeline.} Our training framework jointly optimizes a language modeling loss and a contrastive loss. To mitigate the information leakage problem discussed in \Cref{sec:info-leak}, the model generates rationales on-the-fly during training, ensuring that the embeddings used for contrastive supervision remain independent of oracle rationales and lead to better representation quality.}
    \label{fig:pipeline}
    \vspace{-0.5cm}
\end{figure*}

\subsection{Multimodal Retrieval}
Conventional multimodal retrieval primarily focuses on cross-modal settings such as text-to-image retrieval~\cite{flickr30k, coco, onoe2024docci, sharegpt4v, li2024omnicorpus} and text-to-video retrieval~\cite{chen2011collecting, xu2016msr}, which evaluate how well embeddings align across modalities. Classical approaches typically adopt dual-encoder architectures, e.g. CLIP~\cite{radford2021learning}, EVA-CLIP~\cite{sun2023eva}, and SigLIP~\cite{zhai2023sigmoid}, optimized purely for contrastive matching. More recently, the need for retrieval over composed multimodal inputs has emerged~\cite{wei2024uniir, jiang2025vlmvec}, demanding embeddings that can integrate multiple sources of information. As MLLMs can naturally consume multi-modal instructions and project them into a unified embedding space, they are now increasingly explored as a promising backbone for these composed multimodal retrieval scenarios.

\subsection{Multimodal Large Language Models as Embedding Models}

A recent line of work studies MLLMs as embedding encoders.  Motivated by the success of LLM embeddings~\cite{behnamghader2024llmvec, lee2025nvembed, jiang2024scaling}, E5-V~\cite{jiang2024e5} first showed that with proper prompting, MLLMs can compress semantic information into the final token and begin to close the modality gap.  Subsequent work~\cite{liu2025lamra, zhang2025gme, kong2025unite, lin2025mmembed, thirukovalluru2025b3} further reports that MLLMs can outperform conventional vision–language dual-encoder models on several multimodal retrieval benchmarks, suggesting that generative multimodal architectures have the potential to provide embedding spaces competitive with existing contrastively trained backbones.  Moreover, because MLLMs can naturally accept interleaved multimodal inputs, they support a broader spectrum of input formats than standard dual-tower models, further strengthening their promise as general-purpose multimodal embedders.

However, existing methods do not preserve the generative reasoning capabilities of MLLMs when extracting embeddings. Some prior work, such as CAFe~\cite{yu2025cafe}, explicitly models a trade-off between embedding discriminability and generative ability, optimizing the two objectives separately. VladVA~\cite{ouali2025vladva} also introduces a next-token prediction loss to enhance embedding quality, but the model itself does not perform any reasoning or generate intermediate rationales. In contrast, we treat rationale generation as a useful test-time computation and deliberately retain it, leveraging reasoning as a mechanism to improve discriminative embedding quality.

\subsection{Model Reasoning}
A parallel line of research studies explicit problem-solving procedures to enhance the reasoning ability of (M)LLMs. Chain-of-Thought (CoT) prompting~\cite{wei2022chain} is a key breakthrough in this direction: it improves inference by decomposing a query into multiple intermediate steps. More recent models such as DeepSeek-R1~\cite{guo2025deepseek} employ explicit rationale generation as a structured latent computation before producing the final answer. These ideas have also been transferred into multimodal settings~\cite{shao2024visual, huang2025vision}, suggesting that explicit reasoning may generally help MLLMs produce more accurate outputs. Our work builds on this observation but shifts the focus toward representation learning: instead of generating final answers, we investigate whether explicit reasoning can improve the embeddings extracted from MLLMs for retrieval tasks.

\section{Method}

We propose a training framework that integrates autoregressive rationale generation with contrastive representation learning (Figure~\ref{fig:pipeline}). We couple reasoning ability and embedding ability for the purpose of improving embedding quality. Concretely, we curate ground-truth rationales and apply supervised fine-tuning with an autoregressive loss to teach the MLLM to generate structured rationales. In parallel, we apply a contrastive loss to train the model to produce discriminative embeddings. To prevent information leakage (i.e., the embedding trivially aligning with the target through the oracle rationale), we compute contrastive targets based on on-the-fly generated rationales, rather than ground-truth oracle rationales. After fine-tuning on the query-target paired dataset with our completed rationale supervision, the resulting model can perform explicit reasoning before producing the final embedding, enabling reasoning guided embeddings for multimodal retrieval.

In the following section, we first define the problem setting and motivate the requirement for reasoning guided embeddings. We then describe the data preparation and oracle rationale construction, followed by our unified training objectives. Finally, we analyze the information leakage issue that arises if oracle rationales are exposed to the contrastive objective, and present our solution based on on-the-fly rationale generation.

\subsection{Preliminary}
The most common paradigm for using MLLMs as embedding models is: given an input $x$ composed of arbitrary modality compositions, with proper prompting or fine-tuning, the hidden state $h_x$ of the last token from the final transformer layer output is taken as a semantically rich embedding. We note this by:
\begin{align}
    h_x = \Theta(x),
\end{align}
where $\Theta$ stands for the MLLM used to extract respective embeddings.

To make such embeddings discriminative, contrastive learning is typically applied, and InfoNCE is the dominant objective. Formally, given a dataset of query–target pairs $(q, t)$, the InfoNCE~\cite{he2020momentum} loss is:
\begin{align}
    \mathcal{L}_{\text{InfoNCE}} = -\log \frac{\phi(h_q, h_t^+)}{\phi(h_q, h_t^+) + \sum_{t^-\in \mathbb{N}}\left(\phi(h_q, h_t^-) \right)}, \label{eq:info-nce}
\end{align}
where $h_q$ is the query embedding, $h_t^+$ is its positive target, $\mathbb{N}$ denotes a set of negatives, and $\phi(\cdot,\cdot)$ denotes the temperature-scaled cosine similarity function as follows:
\begin{align}
    \phi(h_q, h_t) = \exp \left(\frac{1}{\tau} \cos(h_q, h_t)\right),
\end{align}
where $\tau$ denotes the temperature hyper-parameter. In multimodal retrieval, the objective is to ensure that the query embedding $h_q$ is closer to its matched target embedding $h_t$ than to any negative target, i.e., representation quality is directly reflected by recall and ranking correctness.

However, we argue that extracting $h_q$ directly from the raw query is suboptimal. Instead, we allow the MLLM to first produce a rationale $r$ conditioned on the query:
\begin{align}
    r \sim p_{\text{MLLM}}(r \mid q).
\end{align} 
After generating the rationale $r$, the embedding extracted after this reasoning step, $h_{q, r}$, is computed on a semantically richer latent state. In this view, the resulting embedding $h_{q, r}$ is expected to carry a more informative signal for distinguishing the correct target among candidates, leading to better retrieval performance.

\subsection{Oracle Rationale Generation}
\label{sec:oracle-dataset}
To enable the MLLM to generate rationales that effectively enhance embedding quality, we first provide it with supervised examples of correct reasoning. To this end, we construct a set of \textit{oracle rationales} generated with access to both the query and its ground-truth target. Based on the existing MMEB training samples, we produce an oracle rationale $r_o$ for each query–target pair $(q, t)$, explaining why the given target correctly matches the query. These oracle rationales serve as supervised language modeling targets in our joint training. Importantly, this process introduces no external data; we simply augment each existing query–target pair with its corresponding oracle rationale.

\begin{figure}[t]
    \centering
    \includegraphics[width=0.8\linewidth]{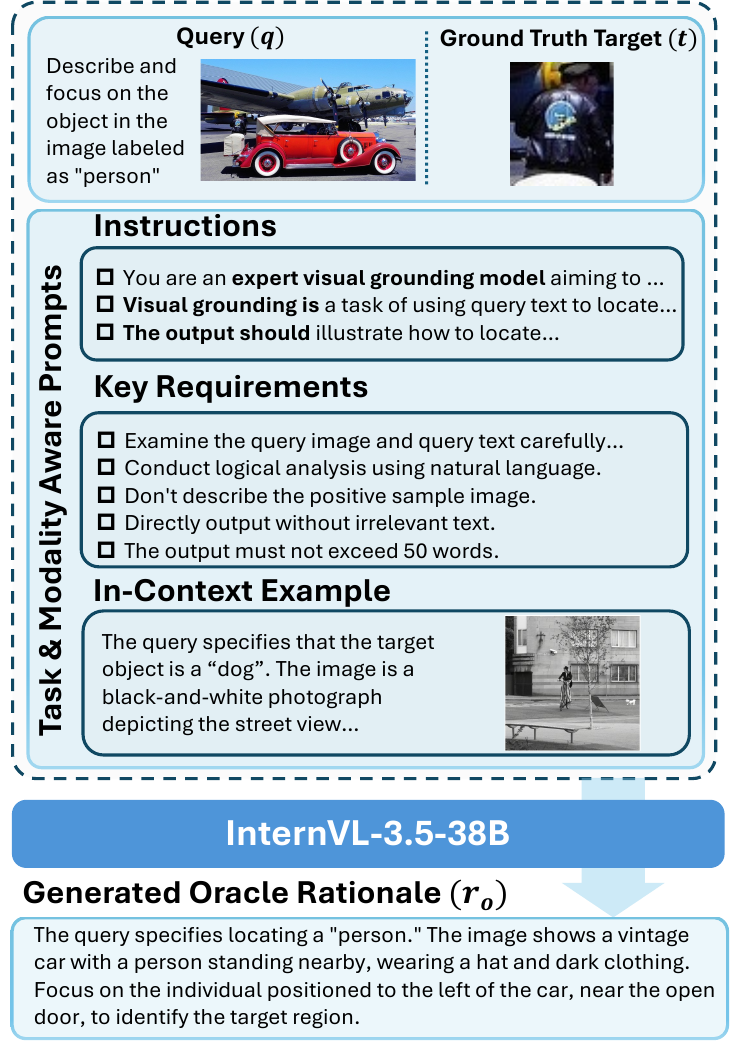}
    \caption{\textbf{Our dataset curation pipeline.} We provide both the query $q$ and its ground-truth target $t$ to an MLLM rationale generator, along with task- and modality-aware prompts designed to produce well-formatted oracle rationales $r_o$. The illustration is selected from the text-image to image grounding task in MMEB.}
    \label{fig:dataset}
    \vspace{-0.5cm}
\end{figure}

As illustrated in \Cref{fig:dataset}, for each downstream task, namely, classification, VQA, retrieval, and visual grounding in MMEB, we manually designed prompts tailored to different modality combinations. Each prompt begins with an instruction that explicitly defines the model's expert role, followed by a concise task explanation clarifying the reasoning goal. Furthermore, we specify clear formatting requirements for the model’s output. The model is instructed to generate an oracle rationale through logical analysis expressed in natural language, while carefully examining both the query image and text via multi-grained description, including general description, object-level details, and task-specific brainstorming based on its world knowledge. At the end of the prompt, an example was attached. Each prompt concludes with an illustrative example, and the expected output is a coherent rationale explaining how to accomplish the given task.


We adopted Intern3.5-38B~\cite{wang2025internvl3_5}, a powerful open-source deployable MLLM, as our oracle rationale generator. After producing oracle rationales for all samples from MMEB, we used the augmented dataset for subsequent supervised fine-tuning.

\subsection{Model Inference Setting}
\label{sec:model-inference}
\textbf{Different inference modes.} At inference time, we want the model to support two modes:(1) directly extract an embedding when no reasoning is needed, and (2) generate rationales when reasoning is beneficial. 

Concretely, if we pre-fill the input with an \texttt{<emb>} token at the final position, the model immediately maps the multimodal input into this token, and we use its final-layer hidden state as the embedding.
If we do not pre-fill \texttt{<emb>}, the model instead enters a generative reasoning mode: it first produces a rationale, terminates at a self-generated \texttt{<emb>}, and then, in order to take the hidden state of this \texttt{<emb>} token, after one more forward step, reusing the KV cache, as the final embedding. This allows a unified model to perform direct pooling or reasoning-before-pooling, without switching architectures.

In practice, we enable reasoning on the query side (which typically requires deeper interpretation), while candidates produce embeddings directly by pre-filling the \texttt{<emb>} token.  Formally:
\begin{align}
    h_{q, r_{\text{self}}} &= \Theta_{RGE}(q, r_{\text{self}}), \\
    h_t &= \Theta_{RGE}(t),
\end{align}
where $(q, r_{\text{self}}, t)$ stand for query, self-generated rationale, and target, respectively. $\Theta_{RGE}$ stands for the RGE model trained under our framework.

\noindent \textbf{Special token setting.} In prior MLLM embedding works, the semantic representation is often taken from the final token embedding (e.g., the \texttt{<eos>} token) at the last hidden layer. In our setting, since we jointly optimize language modeling (for rationale generation) and contrastive alignment (for embedding quality), we do not want any existing token in the vocabulary to simultaneously serve both roles. Therefore, we introduce a dedicated special token \texttt{<emb>} and explicitly use its final-layer hidden state as the pooled embedding. Furthermore, since no text is expected to appear after this token, \texttt{<emb>} also effectively replaces the role of \texttt{<eos>} in this part of the pipeline, allowing the original \texttt{<eos>} to remain purely for language modeling rather than representation pooling.

\begin{table}[t]
\centering
\caption{\textbf{Analysis of information leakage problem.}
We train from scratch on a small subset of MMEB image–text pairs (MSCOCO, VisualNews) with three contrastive supervision configurations: (1) $q \rightarrow t$ (baseline, no rationale), (2) $(q,r_o)\rightarrow t$ (oracle rationale given), and (3) $(q,r_{\text{self}})\rightarrow t$ (on-the-fly self-generated rationale).  The table reports MMEB evaluation results.
}
\label{tab:poc}
\resizebox{0.8\linewidth}{!}{
\begin{tabular}{ccccc} 
\toprule
\multirow{2}{*}{\shortstack{Contrastive\\Supervision Pair}} & \multicolumn{2}{c}{VisualNews} & \multicolumn{2}{c}{MSCOCO}  \\ 
\cmidrule{2-5}
                                 & T2I & I2T                     & T2I & I2T                   \\ 
\midrule
$q \rightarrow t$ & 67.3	& 68.2	& 67.9	& 64.7                    \\
$(q, r_o)\rightarrow t$ & 46.2 &	59.4 &	56.5 &	55.1                      \\
$(q, r_{\text{self}})\rightarrow t$  & \textbf{71.5} &	\textbf{70.8} & \textbf{68.0}	& \textbf{65.8}            \\
\bottomrule
\end{tabular}
}
\vspace{-0.5cm}
\end{table}

\subsection{Joint Training}
\label{sec:unified-training}

\noindent \textbf{Next-Token prediction loss.} Following the standard practice in autoregressive models~\cite{radford2018improving}, we train the MLLM to predict the next token in the rationale sequence.  This encourages the model to internalize a coherent reasoning path.  The language modeling loss is:
\begin{align}
    \mathcal{L}_{\text{LM}}
= - \sum_{i=1}^{N} \log p(y_i \mid y_{<i}, x),
\end{align}
where $x$ is the multimodal input, $y_i$ is the rationale token at position $t$, $N$ is the number of tokens the model is trained to predict, and $p(\cdot)$ denotes the model’s predictive distribution over the vocabulary.

During training, we imitate the two inference behaviors described in \Cref{sec:model-inference}: queries are trained to generate rationales, while candidates are trained to directly produce embeddings without reasoning. For symmetry and stability, we apply the language modeling objective on both sides, but with different target tokens: the query side predicts the rationale tokens, and the candidate side predicts a terminating \texttt{<emb>} token. As illustrated in \Cref{fig:pipeline}, the supervised LM tokens are explicitly marked: multiple rationale tokens on the query side, and a single terminating \texttt{<emb>} token on the target side.

Formally, the weighted language loss for both sides is:

\begin{align}
\bar{\mathcal{L}}_{\text{LM}}
&= - \frac{1}{N_q + N_c}
\Bigg[
N_q\sum_{i=1}^{N_q} \log p\!\left(y^{q}_{i} \mid y^{q}_{<i}, x^{q}\right) \notag
\\
& \qquad\qquad\qquad\ \
+ N_t \sum_{i=1}^{N_t} \log p\!\left(y^{t}_{i} \mid y^{t}_{<i}, x^{t}\right)\Bigg] . \label{eq:lm-loss}
\end{align}

In practice, $N_q$ stands for the number of rationale tokens on the query side, and $N_t = 1$ only at the terminal \texttt{<emb>} position on the target side.

\noindent \textbf{Contrastive Alignment on Self-Generated Rationales.} Following the common practice of contrastive learning, we adopt the same InfoNCE loss in \Cref{eq:info-nce}. After extracting the embeddings for both query and target side, the contrastive loss $\mathcal{L}_{con}$ could be formally expressed as: 
\begin{align}
    \mathcal{L}_{con} = -\log \frac{\phi(h_{q, r_{\text{self}}}, h_t^+)}{\phi(h_{q, r_{\text{self}}}, h_t^+) + \sum_{t^-\in \mathbb{N}}\left(\phi(h_{q, r_{\text{self}}}, h_t^-) \right)}. \label{eq:our-info-nce}
\end{align}

In the following section, we explain why we generate the rationale $r_{\text{self}}$ on-the-fly, rather than directly using the hidden state of the \texttt{<emb>} token from the autoregressive training process with the oracle rationale $r_o$.

\subsection{Information Leakage Problem}
\label{sec:info-leak}

In supervised LM training, autoregressive loss is computed after encoding the entire input sequence, including the prefilling context and the supervised rationale tokens. This implies that the hidden state of the final \texttt{<emb>} token on the query side has full access to oracle rationales. If we directly use this hidden state for contrastive alignment, the task becomes trivial: alignment would not measure representation quality but simply exploit information leakage from the oracle rationale.

\begin{figure}[t]
  \centering
  \begin{subfigure}{0.49\linewidth}
    \includegraphics[width=\linewidth]{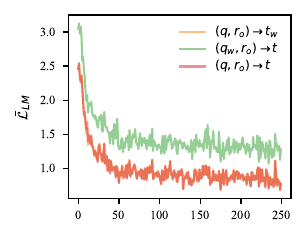}
    \caption{$\mathcal{\bar{L}}_{LM}$ training process.}
  \end{subfigure}
  \hfill
  \begin{subfigure}{0.49\linewidth}
    \includegraphics[width=\linewidth]{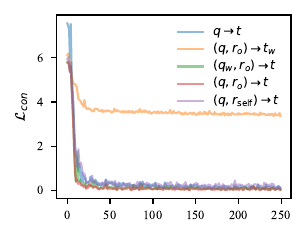}
    \caption{$\mathcal{L}_{con}$ training process.}
  \end{subfigure}
  \caption{\textbf{Retrieval Experiment of Information Leakage.} We conduct diagnostic experiments showing that using the \texttt{<emb>} hidden state from the autoregressive LM loss for contrastive training leads to information leakage. 
When the model is trained on unpaired queries $q_w$ but with correctly paired oracle rationales $r_o$ and targets $t$, namely $(q_w, r_o)\rightarrow t$, the contrastive loss still converges, indicating that the oracle rationale enables the model to take a shortcut rather than learning true query–target alignment.}
\vspace{-0.5cm}
  \label{fig:info-leak}
\end{figure}

We empirically verify this in \Cref{tab:poc} by conducting a controlled retrieval experiment on a small text-image pair training subset of MMEB (MSCOCO, VisualNews). 
Specifically, we train three variants from scratch using identical query target $(q, t)$ pairs and hyper-parameters, but differ only in what representation is used as the contrastive supervision anchor: (1) $q\rightarrow t$, baseline without any rationale; (2) $(q,r_o)\rightarrow t$, taking the \texttt{<emb>} hidden state inside the autoregressive LM forward (thus exposing the oracle rationale $r_o$ tokens to the representation); and (3) $(q,r_{\text{self}})\rightarrow t$, our proposed on-the-fly self-generated rationale before extracting \texttt{<emb>}.  Results in \Cref{tab:poc} show a striking phenomenon: although (2) technically has more (oracle) information than (1), its performance collapses far below the baseline.  In contrast, (3), where the rationale is internally generated rather than given, achieves the best performance across both datasets. This supports our central claim: the hidden state used inside the LM loss is entangled with oracle rationale tokens, which creates an easy shortcut for contrastive training that bypasses the actual query semantics. By instead requiring the model to \emph{self-generate} the rationale, we mitigate this shortcut, and the contrastive loss is forced to align on the correct semantic signal.

We attribute this to the information leakage problem discussed above.  To further validate this, we run a diagnostic experiment with two perturbed triplets $(q,r_o,t)$: (1) $(q, r_o, t_w)$, where $t_w$ is a wrong target for query $q$; and (2) $(q_w, r_o, t)$, where $q_w$ does not match the oracle rationale $r_o$ and the target $t$.

From \Cref{fig:info-leak}, we observe that with triplet $(q,r_o,t_w)$, the autoregressive loss decreases normally because $r_o$ is consistent with $q$. And the contrastive loss does not decrease, as expected, since none of $(q,r_o)$ is aligned with $t_w$. In contrast, with $(q_w,r_o,t)$, the autoregressive loss barely decreases (as $r_o$ does not correspond to $q_w$), but the contrastive loss still decreases surprisingly well, even though $(q_w,t)$ is not a valid pair.  This confirms that if we directly reuse the \texttt{<emb>} hidden state from the autoregressive LM forward pass, the model can take a shortcut in contrastive learning and align $r_o$ with $t$ without relying on the original query $q$.

To prevent this leakage, we remove the ground-truth rationales from the model input for contrastive learning, and let the model generate the rationale on-the-fly at each training step. This ensures that contrastive alignment is conditioned only on the model’s self-produced reasoning. Thus, reasoning quality and discriminative representation learning become directly coupled. 

Therefore, to combine the language modeling loss \Cref{eq:lm-loss} and contrastive loss \Cref{eq:our-info-nce} together, we have:
\begin{align}
    \mathcal{L} = \alpha_{lm} \mathcal{\bar{L}}_{LM} + \alpha_{con} \mathcal{L}_{con}, \label{eq:loss}
\end{align}
where $\alpha_{lm} + \alpha_{con} = 1$. $\alpha_{lm}, \alpha_{con}$ are the coefficient for loss function balancing.

\begin{table*}[t]
\centering
\caption{\textbf{Performance comparison on the MMEB~\cite{jiang2025vlmvec} benchmark.} We highlight the best results in each column in \textbf{bold} and \underline{underline} the second-best results. We report Precision@1 for all models. }
\label{tab:main}
\resizebox{0.9\linewidth}{!}{
\begin{tabular}{ccccccccc} 
\toprule
\multirow{2}{*}{Model}                 & \multirow{2}{*}{MLLM Backbone} & \multicolumn{4}{c}{Per Meta-Task Score}       & \multicolumn{3}{c}{Average Score}  \\ 
\cmidrule(lr){3-6}\cmidrule(lr){7-9}
                                       &                                & Classification & VQA  & Retrieval & Grounding & IND  & OOD  & Overall              \\ 
\midrule
\multicolumn{1}{l}{\textit{Baseline Models}}    &                                &                &      &           &           &      &      &                      \\
CLIP~\cite{radford2021learning}                                   & -                              & 42.8           & 9.1  & 53.0      & 51.8      & 37.1 & 38.7 & 37.8                 \\
SigLIP~\cite{zhai2023sigmoid}                                 & -                              & 40.3           & 8.4  & 31.6      & 59.5      & 32.3 & 38.0 & 34.8                 \\
UniIR~\cite{wei2024uniir}                                  & -                              & 42.1           & 15.0 & 60.1      & 62.2      & 44.7 & 40.4 & 42.8                 \\
Magiclens~\cite{zhang2024magiclens}                              & -                              & 38.8           & 8.3  & 35.4      & 26.0      & 31.0 & 23.7 & 27.8                 \\
\midrule
\multicolumn{1}{l}{\textit{Supervised by MMEB}} &                                &                &      &           &           &      &      &                      \\
VLM2Vec~\cite{jiang2025vlmvec}                                & Qwen2-VL-2B                    & 59.0           & 49.4 & 65.4      & 73.4      & 66.0 & 52.6 & 59.3                 \\
VLM2Vec~\cite{jiang2025vlmvec}                                & Phi-3.5-V-4.2B                 & 54.8           & 54.9 & 62.3      & 79.5      & 66.5 & 52.0 & 60.1                 \\
UNITE~\cite{kong2025unite}                                  & Qwen2-VL-2B                    & 63.2           & 55.9 & 65.4      & 75.6      & 65.8 & 60.1 & 63.3                 \\
MMRet~\cite{zhou2025megapairs}                                  & LLaVA-1.6-7B                   & 56.0           & 57.4 & 69.9      & 83.6      & 68.0 & 59.1 & 64.1                 \\
UniME\cite{gu2025breaking}                                  & Phi-3.5-V-4.2B                 & 54.8           & 55.9 & 64.5      & 81.8      & 68.2 & 52.7 & 64.2                 \\
IDMR~\cite{liu2025idmr}                                   & InternVL2.5-8B                 & 58.3           & 58.6 & 68.7      & 85.6      & 70.5 & 57.9 & 64.9                 \\
LLaVE~\cite{lan2025llave}                                  & Aquila-VL-2B                   & 62.1           & 60.2 & 65.2      & 84.9      & 69.4 & 59.8 & 65.2                 \\
VLM2Vec~\cite{jiang2025vlmvec}                                & Qwen2VL-7B                     & 62.6           & 57.8 & 69.9      & 81.7      & 72.2 & 57.8 & 65.8                 \\
MoCa~\cite{chen2025moca}                                   & Qwen2.5-VL-3B                  & 59.8           & \underline{62.9} & \underline{70.6}      & \underline{88.6}      & \underline{72.3} & 61.5 & 67.5                 \\
B3~\cite{thirukovalluru2025b3}                                     & Qwen2VL-2B                     & \textbf{67.0}           & 61.2 & \textbf{70.9}      & 79.9      & 72.1 & \underline{63.1} & \underline{68.1}                 \\
\hdashline
Baseline (w/o reasoning)                               & Qwen2.5VL-3B                   & 59.5           & 59.9 & 66.8      & 88.0      & 70.6 & 58.4 & 65.2                 \\
\rowcolor[rgb]{0.929,0.902,0.973}
RGE (Ours)                             & Qwen2.5VL-3B                   & \underline{64.4}           & \textbf{67.8} & 70.2      & \textbf{90.1}      & \textbf{75.3} & \textbf{63.7} & \textbf{70.1}                 \\
\bottomrule
\end{tabular}
}
\end{table*}


\section{Experiments}
\subsection{Experiment Settings}

\noindent \textbf{Dataset and Benchmarks.} We evaluate on MMEB~\cite{jiang2025vlmvec}, a multimodal retrieval benchmark comprising 36 datasets across four task types: classification, VQA, retrieval, and grounding. Each dataset is converted into a retrieval setting and evaluated with query–target matching recall. MMEB includes 20 in-domain and 16 out-of-domain tasks, providing a comprehensive evaluation of embedding quality.

\noindent \textbf{Implementation Details.} We adopt Qwen2.5-VL-3B as the backbone MLLM and train the whole model with full fine-tuning. The loss-balancing coefficient is set to $\alpha_{emb}:\alpha_{lm}=10:1$ for better performance. The temperature of the contrastive loss is set to $0.03$. We train on MMEB with oracle rationales that we completed via prompting InternVL3.5-38B~\cite{wang2025internvl3_5}. All experiments are trained for just one epoch following VLM2Vec~\cite{jiang2025vlmvec}. We use a total batch size of 512, with a learning rate of $2\times 10^{-5}$, and set the maximum new-token generation length to 128.

Before any downstream training, we perform a cold start stage~\cite{guo2025deepseek} on a 160K–instance subset using only the language modeling loss, such that the MLLM internalizes the desired output format. This cold-start checkpoint serves as the initialization for all experiments in this paper. For ablations, we also stay within this 160K subset for efficiency.

\subsection{Comparison with Other Methods}
For prior models that are not trained on MMEB, we include CLIP~\cite{radford2021learning}, SigLIP~\cite{zhai2023sigmoid}, UniIR~\cite{wei2024uniir}, MagicLens~\cite{zhang2024magiclens}, and E5-V~\cite{jiang2024e5}, following the evaluation setup adopted in VLM2Vec-v1~\cite{jiang2025vlmvec}. We also include representative models direclty supervised by MMEB, including VLM2Vec~\cite{jiang2025vlmvec}, UNITE~\cite{kong2025unite}, MMRet~\cite{zhou2025megapairs}, UniME~\cite{gu2025breaking}, IDMR~\cite{liu2025idmr}, LLaVE~\cite{lan2025llave}, MoCa~\cite{chen2025moca} and B3~\cite{thirukovalluru2025b3}.

\begin{table}
\centering
\caption{\textbf{Ablation on loss balancing coefficient $\alpha_{LM}$ and $\alpha_{con}$ defined in \Cref{eq:loss}.}}
\label{tab:loss-coefficient}
\resizebox{\linewidth}{!}{
\begin{tabular}{cccccc} 
\toprule
\multirow{2}{*}{$\alpha_{LM}: \alpha_{con}$} & \multicolumn{4}{c}{Per Meta-Task Score} & \multirow{2}{*}{Overall}  \\ 
\cmidrule{2-5}
                                                     & CLS  & VQA  & Retrieval & Grounding     &                           \\ 
\midrule
100:1                                                & 57.0 & 54.1 & 50.0      & 50.7          & 53.1                      \\
20:1                                                 & 59.7 & 61.0 & 58.2      & 71.2          & 60.8                      \\
10:1                                                 & 60.9 & 63.4 & 59.6      & 74.5          & 62.7                      \\
0                                                    & 58.5 & 62.2 & 63.3      & 82.3          & 63.7                      \\
1:1                                                  & 62.5 & 65.8 & 64.1      & 82.9          & 66.2                      \\
1:10                                                 & \textbf{62.9} & \textbf{67.5} & 65.4      & 84.5          & \textbf{67.4}                      \\
1:20                                                 & 62.3 & 66.9 & 65.3      & \textbf{84.8}          & 67.0                      \\
1:100                                                & 62.0 & 67.2 & \textbf{65.8}      & 84.4          & 67.2                      \\
\bottomrule
\end{tabular}
}
\end{table}
\begin{table}[t]
\centering
\caption{\textbf{Ablation on different contrastive supervision pairs.} $q, r_o, t, r_{\text{self}}$ represents query, oracle rationale, target and self-generated rationale respectively, detailed discussed in \Cref{sec:info-leak}. }
\label{tab:self-inference}
\resizebox{\linewidth}{!}{
\begin{tabular}{cccccc} 
\toprule
\multirow{2}{*}{\shortstack{Contrastive\\Supervision Pair}} & \multicolumn{4}{c}{Per Meta-Task Score} & \multirow{2}{*}{Overall}  \\ 
\cmidrule{2-5}
                              & CLS  & VQA  & Retrieval & Grounding     &                           \\ 
\midrule
$q\rightarrow t$                      & 58.5 & 61.8 & 63.3      & 82.3          & 63.7                      \\
$(q, r_o) \rightarrow t$                           & 57.4 & 61.8 & 53.8 & 84.2  & 60.4  \\
$(q, r_{\text{self}}) \rightarrow r$                            & \textbf{62.9} & \textbf{67.5} & \textbf{65.4}      & \textbf{84.5}          & \textbf{67.4}                      \\
\bottomrule
\end{tabular}
}
\vspace{-0.5cm}
\end{table}

Our method achieves the strongest performance among models at a comparable parameter scale; notably, our 2B variant already surpasses several existing 7B models.  To ensure a fair comparison, we also train a baseline model from the same backbone MLLM (Qwen2.5VL~\cite{bai2025qwen25vltechnicalreport}), but without reasoning and directly supervising the final hidden state with a contrastive objective same as \Cref{eq:info-nce}.  Relative to this strongest non-reasoning baseline at the same scale, our method yields substantial improvements on MMEB, demonstrating the effectiveness of reasoning guided embedding (RGE) learning.

{
\renewcommand\tabularxcolumn[1]{m{#1}} 

\begin{table*}[t]
\caption{\textbf{Qualitative comparison between ours and the non-reasoning baseline.} The visualization examples are drawn from MMEB~\cite{jiang2025vlmvec} subsets (ImageNet-A~\cite{hendrycks2021nae}, A-OKVQA~\cite{schwenk2022okvqa}, MSCOCO~\cite{coco}, and RefCOCO~\cite{kazemzadeh-etal-2014-referitgame}), representing four different task types. Additional visualizations are provided in the appendix section \ref{sec:more-vis}.}
\vspace{-0.3cm}
\label{tab:visualization}
\centering
\resizebox{0.8\linewidth}{!}{
\begin{tabularx}{\textwidth}{
  >{\centering\arraybackslash}m{3.5cm}   
  >{\centering\arraybackslash}m{2.5cm}   
  >{\centering\arraybackslash}X          
  >{\centering\arraybackslash}m{1.5cm}   
  >{\centering\arraybackslash}m{1.5cm}   
}
\toprule
Input Image & Input Instruction & Reasoning (Ours) & Retrieval Results (Ours) & Retrieval Results (Baseline) \\
\midrule

\cellimg[\linewidth]{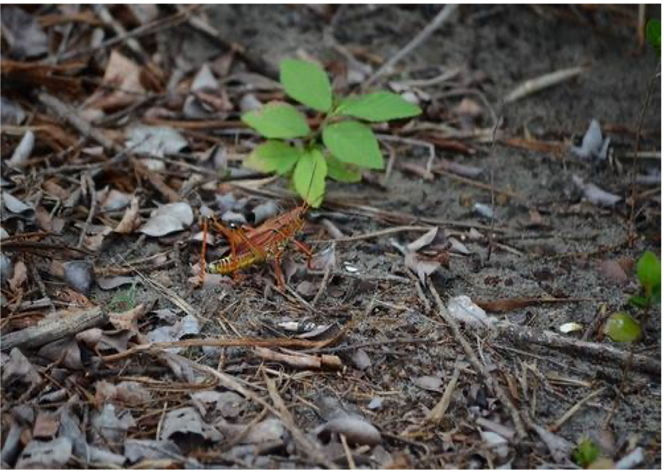}
& Identify the category of the main object in the given image for classification
& The image shows an insect with elongated body and legs, resembling a grasshopper. Its vibrant orange coloration and prominent antennae match typical grasshopper features. The environment, with dry leaves and sparse vegetation, aligns with grasshopper habitats, confirming it as a grasshopper.~
& \makecell{grasshopper,\\hopper}
& cricket \\
\midrule

\cellimg[\linewidth]{}
& Represent the given image with the following question: What are the boats shaped like?
& The image shows a dock with colorful boats resembling small cars, each with a canopy and seating area. These features align with the description of ``cars," distinguishing them from typical boat shapes, confirming the answer ``cars" to the question about the boats' shape.~
& cars
& boat \\
\midrule

\cellimg[\linewidth]{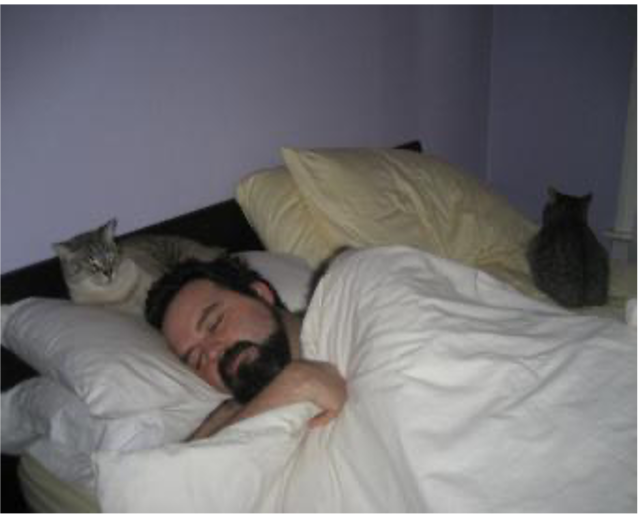}
& Describe and represent the above everyday image for a corresponding description retrieval.
& The image shows a man sleeping in bed with two cats beside him. The bed has white sheets and pillows, and the cats are positioned on either side of the man. The scene suggests a cozy, relaxed atmosphere, matching descriptions of a man sleeping with cats nearby.~
& A man that is laying down underneath a cat.
& Two cats sleeping on top of a bed next to pillows.~ \\
\midrule

\cellimg[\linewidth]{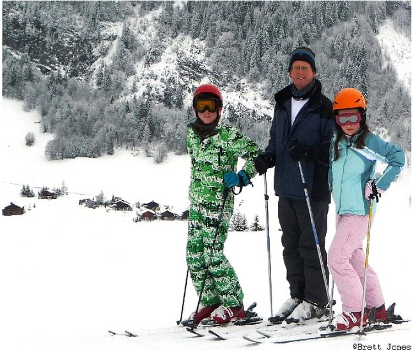}
& Describe and focus on the object in the image that follows the following language expressions: ``man in center"
& The query specifies locating the ``man in center." The image shows three skiers, with the man positioned between two children. He wears a dark jacket and pants, holding ski poles. Focus on the central figure, identifiable by his attire and central placement among the skiers.~
& \cellimg[0.6\linewidth]{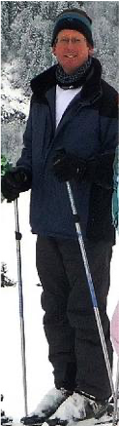}
& \cellimg[0.6\linewidth]{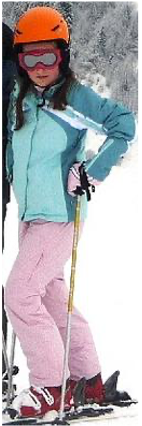} \\
\bottomrule
\end{tabularx}
}
\vspace{-0.5cm}
\end{table*}
}

\subsection{Ablation Study}

\noindent\textbf{Loss balancing coefficient.}
We analyze the loss weighting between the LM loss and the contrastive loss in \Cref{tab:loss-coefficient}. We sweep $\alpha_{LM}:\alpha_{con} = 1/100, 1/20, 1/10, 1, 20, 100$ and also include the baseline without LM supervision $(\alpha_{LM}:\alpha_{con} = 0)$. We observe that increasing the contrastive weight initially improves MMEB performance, indicating stronger representation quality. However, excessively large contrastive weight eventually hurts performance, implying that the LM objective is not merely auxiliary, but contributes to extracting better embeddings in our setting. We therefore adopt $\alpha_{LM}:\alpha_{con}=1:10$ as the default configuration for all main results.

\noindent \textbf{Self-generation contrastive loss.}  
As discussed in \Cref{sec:info-leak}, directly taking the \texttt{<emb>} hidden state inside the autoregressive LM forward pass for contrastive supervision leads to information leakage: the model can shortcut by aligning the oracle rationale with the target and ignore the original query.  \Cref{tab:self-inference} confirms this on the full MMEB evaluation set.  In contrast, using on-the-fly self-generated rationales for contrastive learning removes this shortcut and restores the intended dependence on the query.


\noindent \textbf{Reasoning effect.}  
After fine-tuning, our model supports two inference modes as described in \Cref{sec:unified-training}.  A natural question is: Does the reasoning path actually matter at inference time?  To verify this, we disable reasoning and directly extract embeddings for both query and candidate by pre-filling \texttt{<emb>} as the last token. As shown in \Cref{tab:reason}, disabling reasoning leads to a performance drop, indicating that reasoning at inference time is indeed beneficial.

Interestingly, even under this reasoning-free setting, the model still outperforms the non-reasoning baseline.  This suggests that training with rationales not only improves performance when reasoning is enabled at test time, but also yields generally better embeddings even without reasoning.

\begin{table}
\centering

\caption{\textbf{Ablation on reasoning usage at inference.}}
\label{tab:reason}
\vspace{-0.3cm}
\resizebox{\linewidth}{!}{
\begin{tabular}{cccccc} 
\toprule
\multirow{2}{*}{w/ Reasoning} & \multicolumn{4}{c}{Per Meta-Task Score} & \multirow{2}{*}{Overall}  \\ 
\cmidrule{2-5}
                              & CLS  & VQA  & Retrieval & Grounding     &                           \\ 
\midrule
Baseline                      & 58.5 & 61.8 & 63.3      & 82.3          & 63.7                      \\
\ding{55}                            & 62.8 & 64.3 & 63.3      & 83.7          & 65.7                      \\
\ding{51}                            & \textbf{62.9} & \textbf{67.5} & \textbf{65.4}      & \textbf{84.5}          & \textbf{67.4}                      \\
\bottomrule
\end{tabular}
}
\vspace{-0.5cm}
\end{table}

\vspace{-0.1cm}
\subsection{Case Study}
\vspace{-0.1cm}

To illustrate how reasoning enhances representation quality, we visualize examples from four representative MMEB tasks~\cite{jiang2025vlmvec}: classification, VQA, retrieval, and grounding, using samples from ImageNet-A~\cite{hendrycks2021nae}, A-OKVQA~\cite{schwenk2022okvqa}, MSCOCO~\cite{coco}, and RefCOCO~\cite{kazemzadeh-etal-2014-referitgame} subsets, respectively. For each task, we compare our model with the non-reasoning baseline in \Cref{tab:visualization}.

Across tasks, we observe a consistent trend: the non-reasoning baseline often relies on shallow visual or lexical cues, producing plausible but incorrect matches. In contrast, our model uses world knowledge and explicit reasoning to clarify fine-grained categories, multi-step attribute relations, and referent disambiguation. This leads to embeddings that better reflect intended semantics and improve retrieval and classification accuracy.

These observations support our key insight: reasoning is not a post-hoc explanation, but an essential test-time compute process that directly shapes how MLLMs form meaningful and discriminative multimodal embedding.

\vspace{-0.25cm}
\section{Conclusion}
\vspace{-0.2cm}
We propose Reasoning Guided Embeddings (RGE), a simple framework that explicitly integrates reasoning to improve multimodal embeddings. By preserving the generative rationale process of MLLMs and incorporating it into contrastive training, RGE enables reasoning-guided embedding extraction that captures richer contextual inference signals. By avoiding oracle rationale information leakage and leveraging self-generated rationales as the alignment target, we directly connect reasoning quality with representation discriminability. Experiments on the MMEB benchmark demonstrate substantial improvements over a non-reasoning baseline, surpassing both similarly sized and even some larger models. These results highlight that reasoning is not merely an explanation for answer generation, but a key mechanism for enhancing multimodal embeddings.


{
    \small
    \bibliographystyle{ieeenat_fullname}
    \bibliography{main}
}

\clearpage
\setcounter{page}{1}
\maketitlesupplementary

\begin{table*}[!h]
\centering
\caption{\textbf{Performance comparison on the Flickr30K~\cite{flickr30k} and MSCOCO~\cite{coco} image-text retrieval benchmark.} We highlight the best results in each column in \textbf{bold} and \underline{underline} the second-best results. We also include ablation results for our RGE model that apply reasoning to the query and/or the target, and highlight these variants using a purple background. }
\label{tab:flickr}
\resizebox{\linewidth}{!}{
\begin{tabular}{cccccccccccccc} 
\toprule
\multirow{3}{*}{Model Name}                                  & \multirow{3}{*}{Model Backbone}                                & \multicolumn{6}{c}{Image Retrieval}                      & \multicolumn{6}{c}{Text Retrieval}                        \\ 
\cmidrule{3-14}
                                                             &                                                                & \multicolumn{3}{c}{Flickr30K} & \multicolumn{3}{c}{COCO} & \multicolumn{3}{c}{Flickr30K} & \multicolumn{3}{c}{COCO}  \\ 
\cmidrule(lr){3-5}\cmidrule(lr){6-8}\cmidrule(lr){9-11}\cmidrule(lr){12-14}
                                                             &                                                                & R@1  & R@5  & R@10            & R@1  & R@5  & R@10       & R@1  & R@5  & R@10            & R@1  & R@5  & R@10        \\ 
\midrule
\textit{Baseline Models}                                                         &                                                                &      &      &                 &      &      &            &      &      &                 &      &      &             \\
CLIP-L~\cite{radford2021learning}                                                       & -                                                              & 67.3 & 89.0 & 93.3            & 37.1 & 61.6 & 71.5       & 87.4 & 98.3 & 99.3            & 57.9 & 81.2 & 87.8        \\
EVA-CLIP-5B~\cite{sun2023eva}                                                  & -                                                              & 78.8 & 94.2 & 96.8            & 51.1 & 75.0   & 82.7       & \textbf{93.9} & \textbf{99.4} & \textbf{99.8}            & 68.8 & 87.8 & 92.8        \\
VLM2Vec~\cite{jiang2025vlmvec}                                                      & Qwen2VL-7B                                                     & 69.3 & 90.3 & 94.3            & 47.2 & 72.6 & 81.1       & 87.4 & 97.7 & 99.1            & 60.6 & 82.7 & 89.6        \\
Baseline                                                     & Qwen2.5VL-3B                                                   & 74.9 & 92.5 & 95.7            & 43.8 &	71.0 & 80.8 & 90.6 & 98.7 & \underline{99.5}            &    59.4 & 82.7 & 90.2          \\ 
\midrule
\begin{tabular}[c]{@{}c@{}}Query w/ Reasoning?\end{tabular} & \begin{tabular}[c]{@{}c@{}}Target w/ Reasoning?\end{tabular} &      &      &                 &      &      &            &      &      &                 &      &      &             \\
\rowcolor[rgb]{0.929,0.902,0.973} \ding{55}                                                           & \ding{55}                                                             & 76.0 & 93.0 & 96.1            & 51.1 & 76.5 & 85.0       & 90.4 & 98.6 & 99.4            & 68.3 & 88.8 & 93.7        \\
\rowcolor[rgb]{0.929,0.902,0.973} \ding{55}                                                           & \ding{51}                                                            & 77.7 & \underline{94.4} & 97.1            & 53.2 & 78.6 & \underline{86.8}       & 90.2 & 98.5 & 99.3            & 66.0 & 87.8 & 93.4        \\
\rowcolor[rgb]{0.929,0.902,0.973} \ding{51}                                                          & \ding{55}                                                             & \underline{79.4} & \underline{94.4} & \underline{97.2}            & \underline{53.8} & \underline{78.8} & 86.6       & \underline{92.8} & 99.2 & \underline{99.5}            & \textbf{75.1} & \textbf{91.9} & \textbf{95.9}        \\
\rowcolor[rgb]{0.929,0.902,0.973} \ding{51}                                                          & \ding{51}                                                            & \textbf{80.4} & \textbf{95.2} & \textbf{97.6}            & \textbf{56.3} & \textbf{80.9} & \textbf{88.3}       & 92.1 & \underline{99.3} & \underline{99.5}            & \underline{69.9} & \underline{89.7} & \underline{94.6}        \\
\bottomrule
\end{tabular}
}
\end{table*}

\section{Apply Reasoning to Candidates}
As discussed in \Cref{sec:model-inference}, our \textsc{RGE} model supports two inference modes. Intuitively, we apply reasoning to queries because they are typically expressed as questions or instructions, thus naturally requiring reasoning. However, in principle, reasoning can also be applied to candidates.

To examine this, we conduct experiments on two classic image–text retrieval benchmarks, Flickr30K and COCO, where query–target pairs are symmetric. This serves two purposes: (1) to compare the model’s reasoning behavior when applied to either side of the pair, and (2) to report the performance of \textsc{RGE} on alternative retrieval tasks, providing an additional measure of its generalization ability.

From the results in \Cref{tab:flickr}, we observe that applying reasoning generally improves embedding quality and leads to better performance on image–to–text retrieval. However, for text–to–image retrieval, applying reasoning to both the query and the candidate sometimes underperforms compared with our default configuration (reasoning on the query only). This trend becomes particularly pronounced on the MSCOCO~\cite{coco} retrieval benchmark, where the candidate pool is five times larger than that of Flickr30K~\cite{flickr30k}.

\begin{figure}
    \centering
    \includegraphics[width=0.95\linewidth]{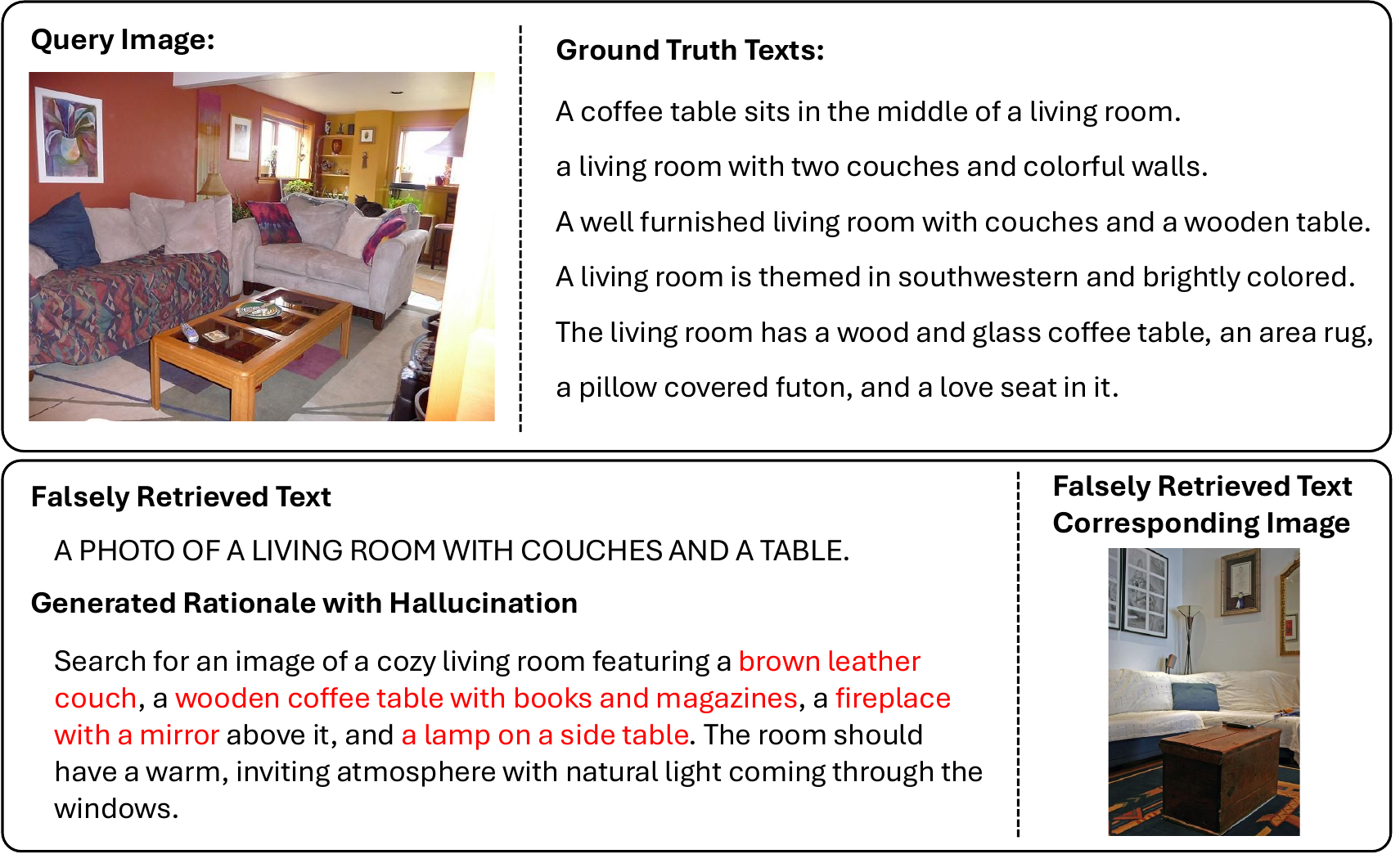}
    \caption{\textbf{Failure Case Demonstration in the MSCOCO~\cite{coco} Text Retrieval Task.} We highlight hallucinated content that does not exist in the falsely retrieved target text in \textcolor{red}{red}. We also include the ground-truth captions for the query image and the image corresponding to the falsely retrieved text.}
    \label{fig:failure}
    \vspace{-0.5cm}
\end{figure}

We suspect that while reasoning enriches contextual information, it may also assign undue emphasis to irrelevant content or even introduce hallucinated details, which is especially harmful on the text side. Text candidates typically contain very limited information compared to image candidates~\cite{schrodi2025two}, making the output embeddings more susceptible to such noise and more easily misled. We demonstrate this in \Cref{fig:failure}, and we can see that reasoning introduces hallucinated information for the text target, which can weaken the overall retrieval performance. Therefore, applying reasoning to candidates, especially those with limited information content in a large gallery, can be less stable and may introduce unnecessary or inaccurate details. 

\vspace{-0.2cm}
\section{Reasoning Failure Case Study}
\label{sec:failure}

Although reasoning enables more successful matches overall, we also observe cases where the reasoning process leads to retrieval errors. \Cref{tab:failure-case} presents three representative types of failure.

The first example shows that an incorrect reasoning path directly results in an incorrect retrieval. The second example illustrates a mismatch in focus: the reasoning emphasizes the content of the whole image, whereas the ground-truth match only depends on the central object, leading to a wrong retrieval result. The third example demonstrates that even with a correct reasoning path, the retrieved answer can still be wrong. This may occur because our RGE model encodes information not only from the generated reasoning path but also from the original inputs. While this mechanism often helps suppress false information in the reasoning, in this case it inadvertently introduces errors into the final retrieval.

{
\renewcommand\tabularxcolumn[1]{m{#1}} 

\begin{table*}[t]
\caption{\textbf{Failure case associated with reasoning.} The visualization examples are drawn from MMEB~\cite{jiang2025vlmvec} subsets (ImageNet~\cite{deng2009imagenet}, MSCOCO~\cite{coco}, and A-OKVQA~\cite{schwenk2022okvqa}). We illustrate three representative types of failure cases: (1) incorrect reasoning leading to incorrect results, (2) reasoning focused on irrelevant content leading to incorrect results, and (3) correct reasoning that nonetheless produces incorrect results.}
\vspace{-0.3cm}
\label{tab:failure-case}
\centering
\resizebox{0.9\linewidth}{!}{
\begin{tabularx}{\textwidth}{
  >{\centering\arraybackslash}m{3.5cm}   
  >{\centering\arraybackslash}m{2.5cm}   
  >{\centering\arraybackslash}X          
  >{\centering\arraybackslash}m{1.5cm}   
  >{\centering\arraybackslash}m{1.5cm}   
}
\toprule
Input Image & Input Instruction & Reasoning (Ours) & Wrong Retrieval Results (Ours) & Ground Truth \\
\midrule

\cellimg[\linewidth]{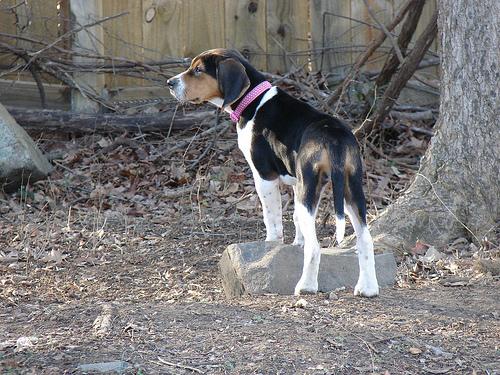}
& Identify the category of the main object in the given image for classification
& The image shows a medium-sized dog with a tricolor coat, standing on a rock. Its short coat, floppy ears, and alert posture align with typical beagle traits. The distinctive black, white, and brown markings confirm it as a beagle.~
& \makecell{beagle}
& Walker hound, Walker foxhound \\
\midrule

\cellimg[\linewidth]{}
& Describe and represent the above everyday image for a corresponding description retrieval.
& The image shows a baseball game in progress, with a pitcher on the mound and players positioned on the field. The setting is an open grassy area with palm trees in the background, suggesting a casual, recreational atmosphere. These elements align with descriptions of a baseball game being played outdoors.~
& A group of people on a field playing baseball.
& A man in a yellow shirt stands in a dirt circle.~ \\
\midrule

\cellimg[\linewidth]{}
& Answer the question based on the given image: What is the profession of the man standing near the net?
& The image shows a tennis player near the net, holding a racket, indicating his role as a professional athlete. The presence of spectators and sponsor logos suggests a competitive match setting, confirming his profession as a tennis player.~
& coach
& athlete \\

\bottomrule
\end{tabularx}
}
\vspace{-0.5cm}
\end{table*}
}

\vspace{-0.1cm}
\section{Oracle Rationale Generation Prompts}

As discussed in \Cref{sec:oracle-dataset}, we designed task- and modality-aware prompts for each training dataset in MMEB~\cite{jiang2025vlmvec}. All prompts we used are displayed in \Cref{tab:rationale_prompts}.

{
\renewcommand{\arraystretch}{0.1}
\setlength{\tabcolsep}{-1pt}
\begin{table*}[t]
\centering
\caption{\textbf{Prompt design example for rationale generation.} We provide a prompt sample for each task.}
\label{tab:rationale_prompts}

\resizebox{0.85\linewidth}{!}{
\begin{tabular}{@{}m{2.5cm}m{13cm}@{}}
\toprule
\makecell{\textbf{Dataset}} & \makecell{\textbf{Prompt}} \\
\midrule

\textbf{ImageNet-1K} & \footnotesize
\textbf{Task:} \newline You are an expert classifier model and need to generate enhanced image classification rationale based on visual description to instruct classification task.  \newline
\textbf{Instructions:} \newline Positive sample is attached at the end (ground truth), use it to verify reasoning but do not mention it explicitly. \newline Examine the query image carefully. \newline Create an enhanced classification rationale that helps classify the query image by using analytical statements, including the query image’s category and attributes.  \newline
\textbf{Requirements:} \newline Conduct logical analysis using natural language. \newline Directly output the generated rationale without irrelevant text. \newline Response must not exceed 50 words. Never use “positive sample/negative sample” in the output.  \newline
\textbf{Example:} \newline The image shows a medium-sized golden dog standing on grass in a park. Its short coat, drooping ears, and sturdy body match retriever traits. Compared to similar breeds, the shorter fur and head confirm it should be classified as a Labrador retriever.\\

\midrule

\textbf{A-OKVQA} & \footnotesize
\textbf{Task:} \newline You are an expert VQA model and need to generate enhanced VQA rationale based on visual description to instruct question answering.  \newline 
\textbf{Instructions:} \newline Positive sample is attached at the end (ground truth) for reasoning verification.\newline  Query text is attached at the end and should guide the rationale. \newline Examine both query image and query text carefully. \newline Create an enhanced rationale that helps answer the question by analytical statement through the query image, including its category and attributes.  \newline 
\textbf{Requirements:} \newline Conduct logical analysis in natural language. \newline Directly output the rationale.\newline  Response must not exceed 50 words. Never use “positive sample.”  \newline 
\textbf{Example:}\newline  The image depicts a kitchen with a table near a window. On the table sits a small bowl holding a shiny red fruit, clearly round and smooth. These attributes identify it as an apple, not an orange, supporting the answer “apple” to the question about the fruit.\\

\midrule

\textbf{MSCOCO\_i2t} & \footnotesize
\textbf{Task:} \newline You are an expert retrieval model and need to generate enhanced image search rationale based on visual description to instruct retrieval task.  \newline 
\textbf{Instructions:} \newline Positive sample is attached at the end (ground truth), use it for verification but not explicit mention. \newline Examine the query image carefully. \newline Create an enhanced rationale that helps retrieve the correct candidate text by analytical statements through the image, including the image’s category and attributes.  \newline 
\textbf{Requirements:} \newline Conduct logical analysis using natural language. \newline Directly output the generated rationale. \newline Response must not exceed 50 words. Never use “positive sample.” \newline  
\textbf{Example:} \newline The scene presents a busy street with a compact silver car at an intersection. Behind it, a furniture shop with bright signage and a cow statue serves as strong cues. Together with traffic lights and buildings, these confirm retrieval should match texts describing this urban street view.\\

\midrule

\textbf{MSCOCO} & \footnotesize
\textbf{Task:} \newline You are an expert visual grounding model aiming to locate the target region in the query image according to the query text.\newline   
\textbf{Instructions:} \newline This is a text grounding task using query text to locate specific regions or objects in an image. \newline Examine query image and query text carefully. \newline Use background knowledge to reason to the correct target region based on visual details. \newline Generate an enhanced rationale that locates the correct region through the text and image, including their attributes.  \newline 
\textbf{Requirements:} \newline Conduct logical analysis using natural language. \newline Directly output the rationale. \newline Response must not exceed 50 words. Never use “positive sample.”  \newline 
\textbf{Example:} \newline The query text specifies the target is a “dog." The image is a black-and-white street view, and the small white dog on the right sidewalk fits the description: medium-sized, short fur, wearing a collar. It stands on a light surface against a dark background—focus on the dog on the right.\\

\bottomrule
\end{tabular}
}
\end{table*}
}

\vspace{-0.1cm}
\section{Additional Case Study}
\label{sec:more-vis}

We provide additional case studies categorized by different tasks: classification (\Cref{tab:cls-visualization}), VQA (\Cref{tab:vqa-visualization}), retrieval (\Cref{tab:ret-t2i-visualization}, \Cref{tab:ret-it2i-visualization}), and visual grounding (\Cref{tab:vgd-visualization}). 

{
\renewcommand\tabularxcolumn[1]{m{#1}} 

\begin{table*}[t]
\caption{\textbf{Additional case studies of classification datasets in the MMEB~\cite{jiang2025vlmvec} benchmark.}}
\vspace{-0.3cm}
\label{tab:cls-visualization}
\centering
\resizebox{0.8\linewidth}{!}{
\begin{tabularx}{\textwidth}{
  >{\centering\arraybackslash}m{1.8cm}   
  >{\centering\arraybackslash}m{3.5cm}   
  >{\centering\arraybackslash}m{2.5cm}   
  >{\centering\arraybackslash}X          
  >{\centering\arraybackslash}m{1.5cm}   
  >{\centering\arraybackslash}m{1.5cm}   
}
\toprule
Dataset & Input Image & Input Instruction & Reasoning (Ours) & Retrieval Results (Ours) & Retrieval Results (Baseline) \\
\midrule

ImageNet-1K & \cellimg[\linewidth]{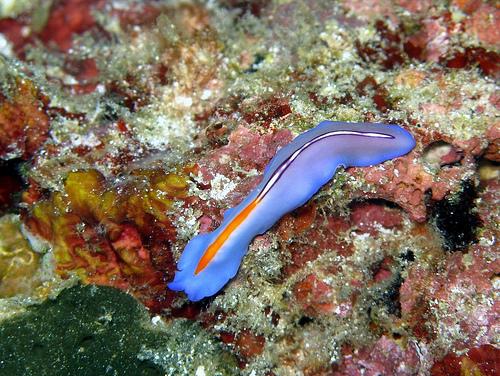}
& Identify the category of the main object in the given image for classification.
& The image depicts a vibrant blue marine creature with distinct orange stripes, resembling a flatworm. Its elongated body and smooth texture align with characteristics of platyhelminths, confirming its classification as a flatworm. The aquatic environment further supports this identification.~
& flatworm, platyhelminth
& sea slug, nudibranch \\
\midrule

ObjectNet & \cellimg[\linewidth]{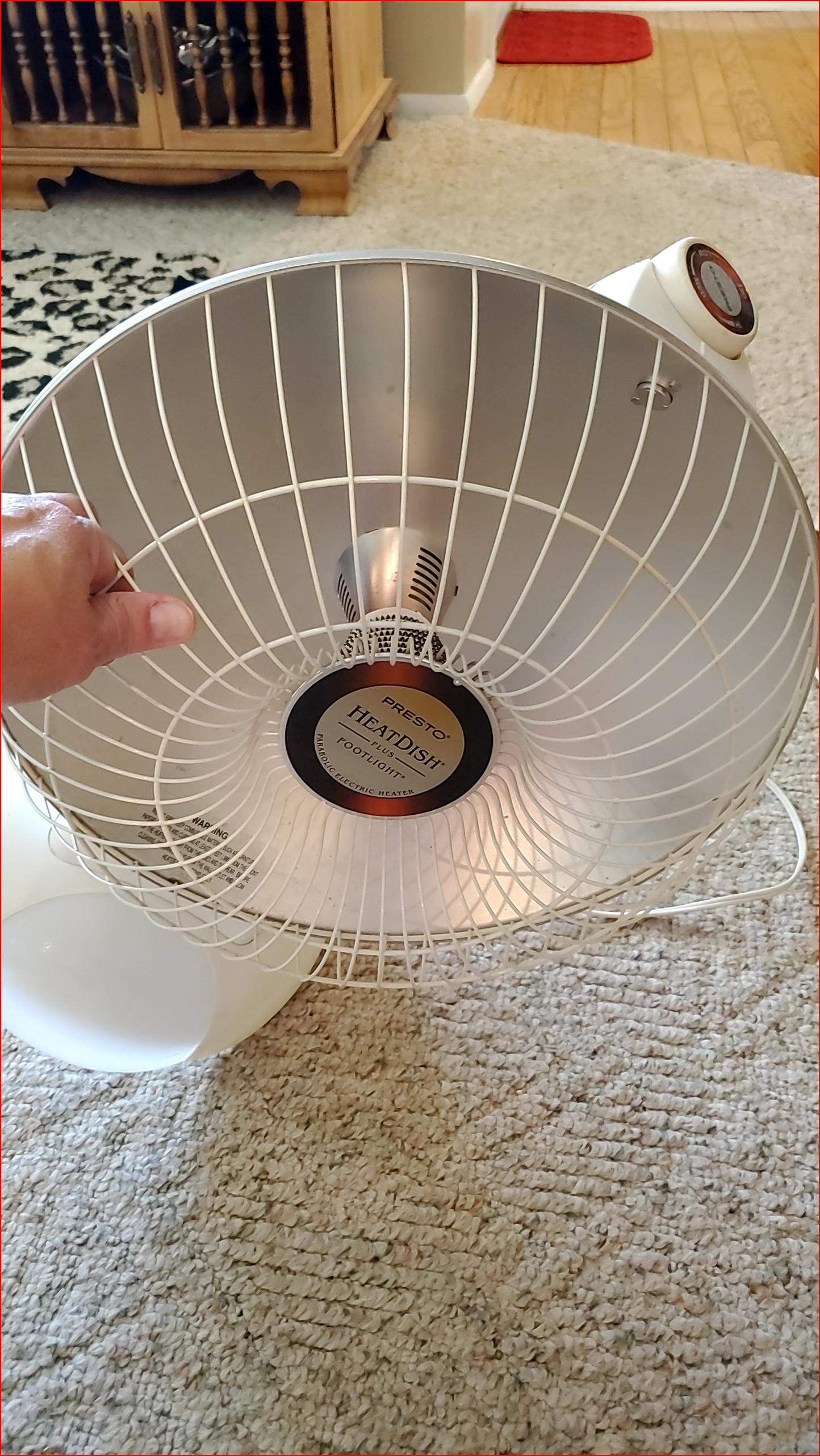}
& Identify the category of the main object in the given image for classification.
& The image shows a white electric heater with a circular design and a visible brand label. Its structure, including the protective grill and control knob, aligns with typical space heaters. The presence of a power cord and its placement on a carpeted floor further confirm it as a space heater.~
& portable heater
& fan \\
\midrule

VOC2007 & \cellimg[\linewidth]{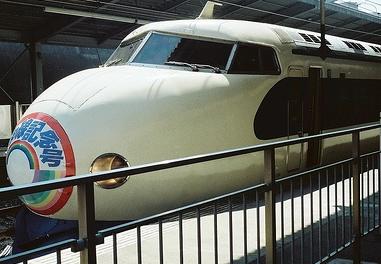}
& Identify the category of the main object in the given image for classification.
& The image shows a sleek, white high-speed train with a streamlined nose and aerodynamic design. Its pointed front and smooth surface are characteristic of bullet trains. The presence of Japanese characters on the side further confirms its classification as a bullet train.~
& train
& aeroplane \\
\midrule

SUN397 & \cellimg[\linewidth]{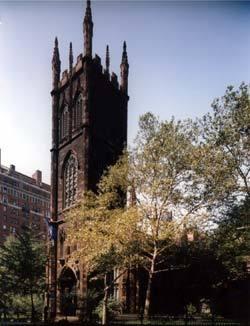}
& Identify the category of the scene in the given image for classification.
& The image depicts a tall, ornate tower with pointed spires and arched windows, characteristic of Gothic architecture. The structure's height and design elements align with typical church towers, confirming it as a church tower.~
& tower
& church outdoor \\

\bottomrule
\end{tabularx}
}
\vspace{-0.5cm}
\end{table*}
}

{
\renewcommand\tabularxcolumn[1]{m{#1}} 

\begin{table*}[t]
\caption{\textbf{Additional case studies of VQA datasets in the MMEB~\cite{jiang2025vlmvec} benchmark.}}
\vspace{-0.3cm}
\label{tab:vqa-visualization}
\centering
\resizebox{0.8\linewidth}{!}{
\begin{tabularx}{\textwidth}{
  >{\centering\arraybackslash}m{1.5cm}   
  >{\centering\arraybackslash}m{3.5cm}   
  >{\centering\arraybackslash}m{2.5cm}   
  >{\centering\arraybackslash}X          
  >{\centering\arraybackslash}m{1.5cm}   
  >{\centering\arraybackslash}m{1.5cm}   
}
\toprule
Dataset & Input Image & Input Instruction & Reasoning (Ours) & Retrieval Results (Ours) & Retrieval Results (Baseline) \\
\midrule

A-OKVQA & \cellimg[\linewidth]{}
& Answer the question based on the given image: What is this umbrella used for?
& The image shows an umbrella with a light source attached, suggesting it's used for photography. The setup indicates it's likely used to diffuse light, enhancing lighting effects in photography, unlike a typical rain umbrella which would not have such attachments.~
& lighting
& rain protection \\
\midrule

ChartQA & \cellimg[\linewidth]{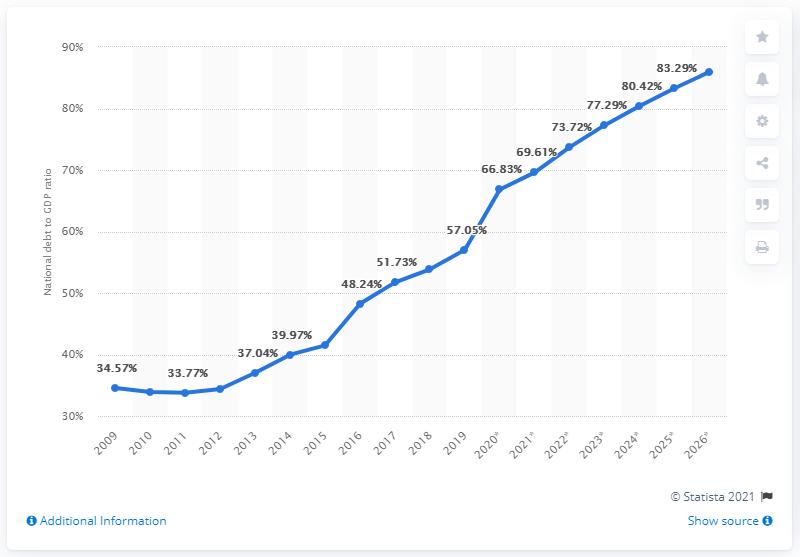}
& Answer a chart-related question based on the given image: What is the difference between the highest and the second highest point?
& The image shows a line graph depicting the national debt to GDP ratio from 2009 to 2026. The highest point is 83.29\% in 2026, and the second highest is 80.42\% in 2025. The difference is 2.87\%, confirming the answer to the query.~
& 2.87
& 0.57 \\
\midrule

OK-VQA & \cellimg[\linewidth]{}
& Answer the question based on the given image: The rounded front part of the plane is also called the what?
& The image shows a vintage propeller plane with a rounded front section. This feature is characteristic of an aircraft's nose, which is distinct from other parts like wings or tail. The rounded shape and position confirm it as the nose, aligning with the query about the plane's front part.~
& nose
& cockpit \\
\midrule

ScienceQA & \cellimg[\linewidth]{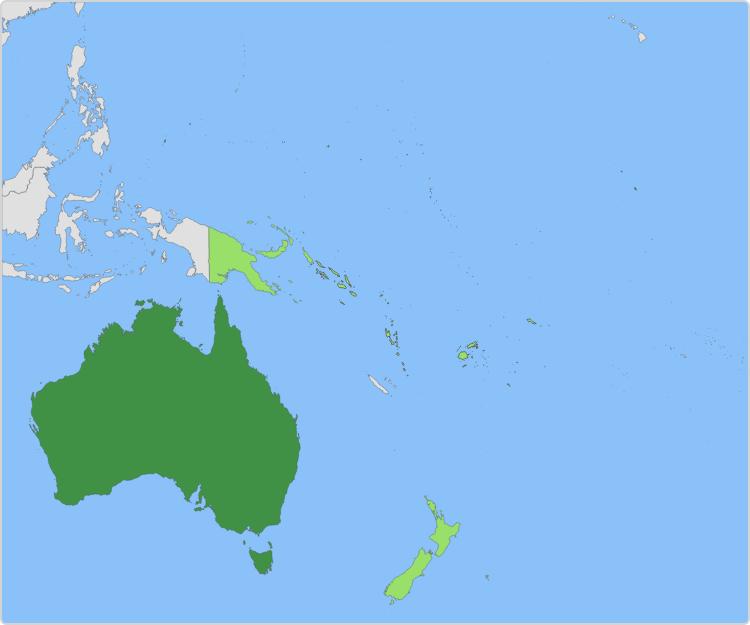}
& Answer a science-related question based on the given image: Which country is highlighted?
& The image shows a map highlighting Australia in green. The highlighted area covers the continent, distinguishing it from other regions like New Zealand and the Pacific Islands. This visual focus confirms Australia as the highlighted country.~
& Australia
& Solomon Islands \\
\midrule

VizWiz & \cellimg[\linewidth]{}
& Answer a question based on the given image: What is this?
& The image shows a white box with an image of a smartphone on it, indicating it contains a mobile device. The design and packaging suggest it is a new smartphone, matching the description of a ``phone."~
& iphone
& ipod nano case \\
\bottomrule
\end{tabularx}
}
\vspace{-0.5cm}
\end{table*}
}

{
\renewcommand\tabularxcolumn[1]{m{#1}} 

\begin{table*}[t]
\caption{\textbf{Additional case studies of retrieval datasets in the MMEB~\cite{jiang2025vlmvec} benchmark.}}
\vspace{-0.3cm}
\label{tab:ret-t2i-visualization}
\centering
\resizebox{0.8\linewidth}{!}{
\begin{tabularx}{\textwidth}{
  >{\centering\arraybackslash}m{1.8cm}   
  >{\centering\arraybackslash}m{4cm}   
  >{\centering\arraybackslash}m{4cm}         
  >{\centering\arraybackslash}m{3cm}   
  >{\centering\arraybackslash}m{3cm}   
}
\toprule
Dataset & Input Text & Reasoning (Ours) & Retrieval Results (Ours) & Retrieval Results (Baseline) \\
\midrule

MSCOCO\_t2i & Describe and represent the following descriptive caption for everyday image retrieval: Two men on motorcycles at a stop light.
& Search for an image showing two men riding motorcycles at a traffic light. The scene should include a busy street with multiple vehicles, including cars and trucks, and a traffic light displaying red. The riders should be wearing helmets and jackets, and the setting should appear to be in an urban environment with buildings and trees in the background.~
& \cellimg[\linewidth]{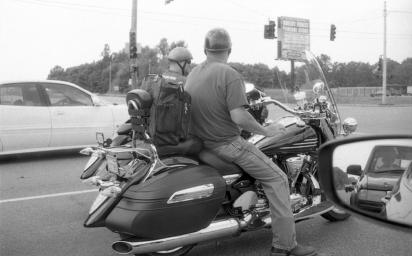}
& \cellimg[\linewidth]{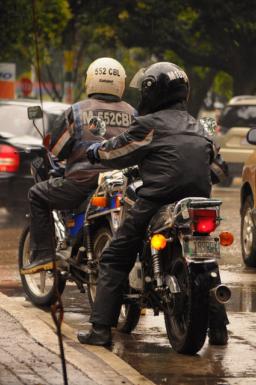}\\
\midrule

VisDial & Describe and represent the following dialogue about an image for image retrieval: Q:is he alone
A:yes
Q:how old is he
A:looks like maybe 26
Q:does he look excited
A:no
Q:does he wear surfing clothes
A:yes
Q:what color is it
A:black, green and red
Q:what color is his board
A:yellow
Q:is it a big wave
A:yes
Q:any fish or shark
A:no
Q:what else are there beside the surfer and his board
A:the sky
Q:is it day time
A:daytime
& A surfer in black, green, and red surfing attire rides a large yellow surfboard on a big wave during daytime, with no other people or marine life visible, under a clear blue sky.~
& \cellimg[\linewidth]{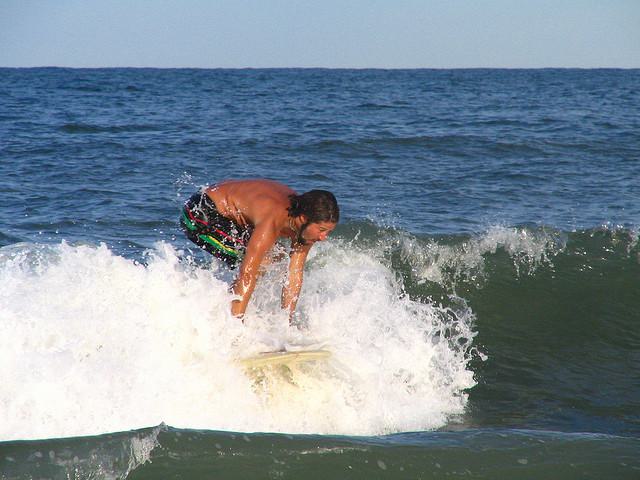}
& \cellimg[\linewidth]{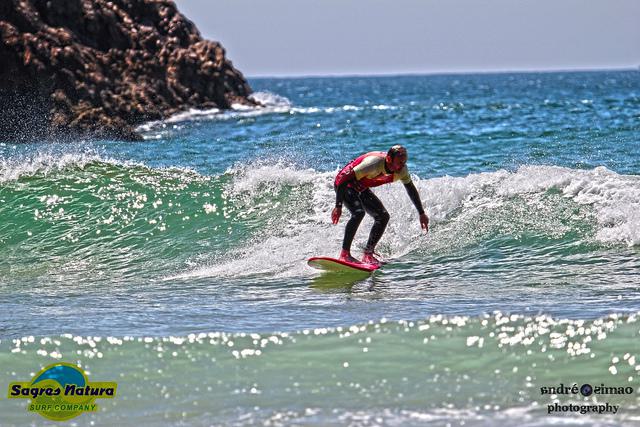} 
\\
\midrule

EDIS & Describe and represent the following provided caption for a news image retrieval: A protester and a police formation on Wednesday at a rally in Athens against austerity.
& Search for an image showing a protest scene in Athens with a large crowd, including a protester holding a sign, and a police formation in the background. The setting should be outdoors with trees and buildings visible, and the atmosphere should reflect a rally against austerity measures.~
& \cellimg[\linewidth]{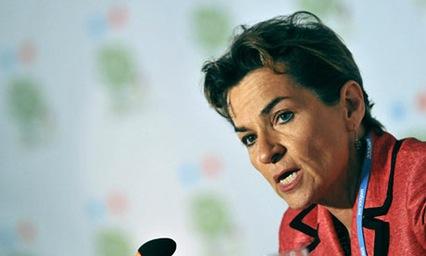} \newline Ahead of Summit, Greece Rushes to Approve New Cuts.
& \cellimg[\linewidth]{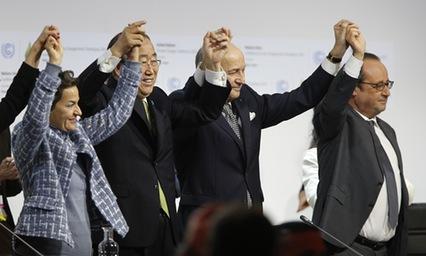} \newline Lucidity has been suspended until further notice in Uruguay.
\\
\bottomrule
\end{tabularx}
}
\vspace{-0.5cm}
\end{table*}
}

{
\renewcommand\tabularxcolumn[1]{m{#1}} 

\begin{table*}[t]
\caption{\textbf{Additional case studies of retrieval datasets in the MMEB~\cite{jiang2025vlmvec} benchmark.}}
\vspace{-0.3cm}
\label{tab:ret-it2i-visualization}
\centering
\resizebox{0.75\linewidth}{!}{
\begin{tabularx}{\textwidth}{
  >{\centering\arraybackslash}m{1.5cm}   
  >{\centering\arraybackslash}m{3cm}   
  >{\centering\arraybackslash}m{2.5cm}   
  >{\centering\arraybackslash}X          
  >{\centering\arraybackslash}m{2cm}   
  >{\centering\arraybackslash}m{2cm}   
}
\toprule
Dataset & Input Image & Input Instruction & Reasoning (Ours) & Retrieval Results (Ours) & Retrieval Results (Baseline) \\
\midrule

CIRR & \cellimg[\linewidth]{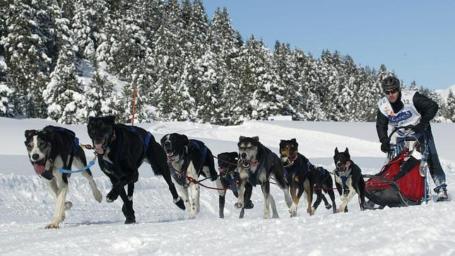}
& Describe and represent the above image and the following described changes about the given image for composed image retrieval: The target photo is the front view of white husky dogs with red booties running ahead.
& The image shows a team of sled dogs pulling a sled through a snowy landscape. The dogs are harnessed and appear to be in motion, with a person guiding them. The background features snow-covered trees under a clear blue sky.

Search Rationale:
To find a similar image, focus on a group of sled dogs in a snowy environment. Ensure the dogs are in motion, wearing harnesses, and the scene includes a snowy landscape with trees. The front view should highlight the white fur and red booties of the dogs.~
& \cellimg[\linewidth]{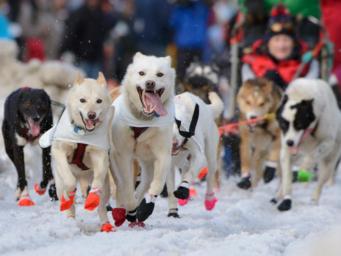}
& \cellimg[\linewidth]{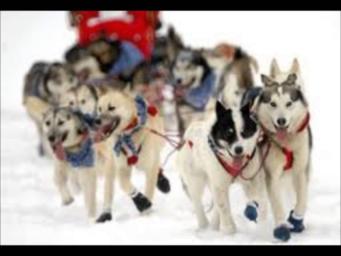} \\
\midrule

NIGHTS & \cellimg[\linewidth]{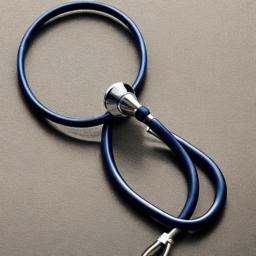}
& Describe and represent the above day-to-day image for a similar image retrieval.
& The image shows a stethoscope with a blue tube and metallic chest piece on a neutral background. The stethoscope is coiled neatly, highlighting its medical design. Look for images featuring a similar stethoscope with a blue tube and metallic components, ensuring the background is simple and unobtrusive.~
& \cellimg[\linewidth]{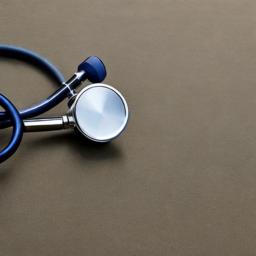}
& \cellimg[\linewidth]{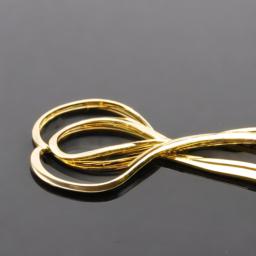} \\
\bottomrule
\end{tabularx}
}
\vspace{-0.5cm}
\end{table*}
}

{
\renewcommand\tabularxcolumn[1]{m{#1}} 

\begin{table*}[t]
\caption{\textbf{Additional case studies of visual grounding datasets in the MMEB~\cite{jiang2025vlmvec} benchmark.}}
\vspace{-0.3cm}
\label{tab:vgd-visualization}
\centering
\resizebox{0.8\linewidth}{!}{
\begin{tabularx}{\textwidth}{
  >{\centering\arraybackslash}m{1.5cm}   
  >{\centering\arraybackslash}m{3cm}   
  >{\centering\arraybackslash}m{2.5cm}   
  >{\centering\arraybackslash}X          
  >{\centering\arraybackslash}m{2cm}   
  >{\centering\arraybackslash}m{2cm}   
}
\toprule
Dataset & Input Image & Input Instruction & Reasoning (Ours) & Retrieval Results (Ours) & Retrieval Results (Baseline) \\
\midrule
RefCOCO & \cellimg[\linewidth]{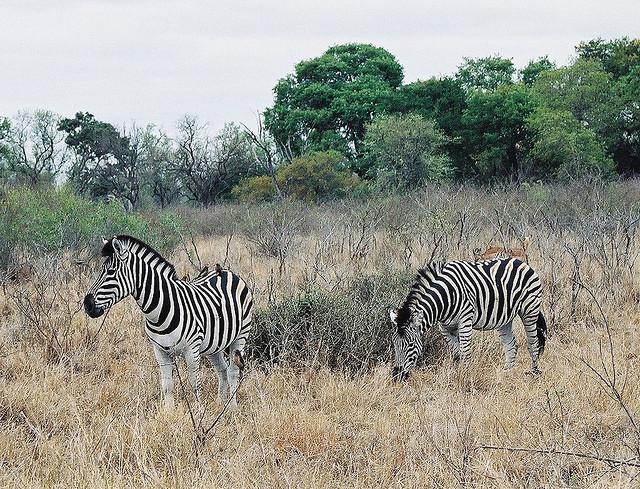}
& Describe and focus on the object in the image that follows the following language expressions. ``right zebra"
& The query specifies locating the "right zebra." The image shows two zebras in a grassy field. Focus on the zebra on the right, which is grazing. It has distinct black and white stripes, standing in dry grass with trees in the background.~
& \cellimg[\linewidth]{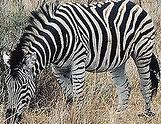}
& \cellimg[\linewidth]{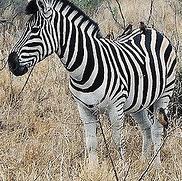} \\
\midrule

MSCOCO & \cellimg[\linewidth]{}
& Describe and focus on the object in the image labeled as ``bicycle".
& The query specifies locating a ``bicycle." The image shows a street scene with motorcycles and bicycles. Focus on the green bicycle in the foreground, positioned prominently on the left side, with visible handlebars and wheels, distinct from the motorcycles behind it.~
& \cellimg[\linewidth]{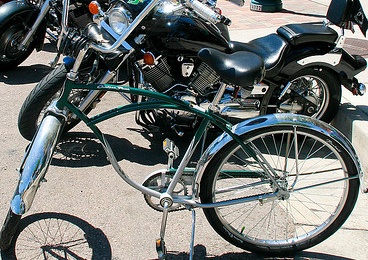}
& \cellimg[\linewidth]{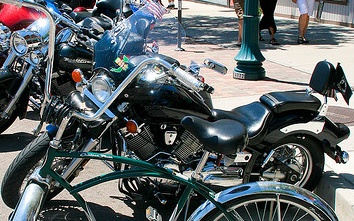} \\
\bottomrule
\end{tabularx}
}
\vspace{-0.5cm}
\end{table*}
}

\vspace{-0.1cm}
\section{Detailed Performance on MMEB}
\Cref{tab:detail} reports the detailed performance of our RGE and baseline models on each subset of the MMEB benchmark~\cite{jiang2025vlmvec}, which comprises 36 tasks across four categories: classification, VQA, retrieval, and visual grounding, including 20 in-domain and 16 out-of-domain datasets.

\begin{table*}
\centering
\caption{\textbf{The detailed performance of the baseline models and our RGE model on MMEB, which contains 20 in-domain datasets and 16 out-of-domain datasets.} We highlight those out-of-domain datasets with a blue background in the table.}
\label{tab:detail}
\resizebox{\linewidth}{!}{
\begin{tabular}{lccccccccc}
\toprule
Dataset                    & CLIP                 & Magiclens            & \begin{tabular}[c]{@{}c@{}}VLM2Vec\\(Phi-3.5-V-4.2B)\end{tabular} & \begin{tabular}[c]{@{}c@{}}UNITE\\(Qwen2-VL-2B)\end{tabular} & \begin{tabular}[c]{@{}c@{}}MMRet\\(LLaVA-1.6-7B)\end{tabular} & \begin{tabular}[c]{@{}c@{}}IDMR\\(InternVL2.5-8B)\end{tabular} & \begin{tabular}[c]{@{}c@{}}MoCa-3B\\(Qwen2.5-VL-3B)\end{tabular} & \begin{tabular}[c]{@{}c@{}}Baseline\\(Qwen2.5-VL-3B)\end{tabular} & \begin{tabular}[c]{@{}c@{}}RAE (Ours)\\(Qwen2.5-VL-3B)\end{tabular}  \\ 
\midrule
\textbf{\textit{Classification (10 tasks)}}  & \multicolumn{1}{l}{} & \multicolumn{1}{l}{} & \multicolumn{1}{l}{}                                              & \multicolumn{1}{l}{}                                         & \multicolumn{1}{l}{}                                          & \multicolumn{1}{l}{}                                           & \multicolumn{1}{l}{}                                             & \multicolumn{1}{l}{}                                              & \multicolumn{1}{l}{}                                                 \\
ImageNet-1K                & 55.8                 & 48.0                 & 65.6                                                              & 77.6                                                         & 58.8                                                          & 70.5                                                           & 75.4                                                             & 72.6                                                              & 80.9                                                                 \\
N24News                    & 34.7                 & 33.7                 & 79.5                                                              & 66.8                                                         & 71.3                                                          & 80.2                                                           & 80.9                                                             & 80.1                                                              & 80.5                                                                 \\
HatefulMemes               & 51.1                 & 49.0                 & 67.1                                                              & 57.4                                                         & 53.7                                                          & 72.9                                                           & 70.6                                                             & 69.2                                                              & 76.4                                                                 \\
VOC2007                    & 50.7                 & 51.6                 & 88.6                                                              & 85.0                                                         & 85.0                                                          & 86.1                                                           & 87.0                                                             & 83.1                                                              & 87.8                                                                 \\
SUN397                     & 43.4                 & 57.0                 & 72.7                                                              & 75.2                                                         & 70.0                                                          & 77.3                                                           & 74.8                                                             & 76.4                                                              & 76.9                                                                 \\
\rowcolor[RGB]{235, 242, 249} Place365                   & 28.5                 & 31.5                 & 42.6                                                              & 41.3                                                         & 43.0                                                          & 44.2                                                           & 38.8                                                             & 42.8                                                              & 45.2                                                                 \\
\rowcolor[RGB]{235, 242, 249} ImageNet-A                 & 25.5                 & 8.0                  & 19.3                                                              & 48.3                                                         & 36.1                                                          & 39.3                                                           & 39.7                                                             & 31.4                                                              & 35.9                                                                 \\
\rowcolor[RGB]{235, 242, 249} ImageNet-R                 & 75.6                 & 70.9                 & 70.2                                                              & 88.8                                                         & 71.6                                                          & 71.6                                                           & 75.4                                                             & 80.2                                                              & 79.0                                                                 \\
\rowcolor[RGB]{235, 242, 249} ObjectNet                  & 43.4                 & 31.6                 & 29.5                                                              & 66.6                                                         & 55.8                                                          & 26.2                                                           & 31.3                                                             & 36.3                                                              & 59.6                                                                 \\
\rowcolor[RGB]{235, 242, 249} Country211                 & 19.2                 & 6.2                  & 13.0                                                              & 24.8                                                         & 14.7                                                          & 14.7                                                           & 24.0                                                             & 22.5                                                              & 21.9                                                                 \\
\rowcolor[RGB]{235, 235, 235} All Classification         & 42.8                 & 38.8                 & 54.8                                                              & 63.2                                                         & 56.0                                                          & 58.3                                                           & 59.8                                                             & 59.5                                                              & 64.4                                                                 \\ 
\midrule
\textbf{\textit{VQA (10 tasks)}}             &                      &                      &                                                                   &                                                              &                                                               &                                                                &                                                                  &                                                                   &                                                                      \\
OK-VQA                     & 7.5                  & 12.7                 & 63.2                                                              & 57.3                                                         & 73.3                                                          & 68.9                                                           & 40.0                                                             & 64.0                                                              & 69.2                                                                 \\
A-OKVQA                    & 3.8                  & 2.9                  & 50.2                                                              & 46.0                                                         & 56.7                                                          & 56.6                                                           & 54.6                                                             & 51.8                                                              & 57.3                                                                 \\
DocVQA                     & 4.0                  & 3.0                  & 78.4                                                              & 87.0                                                         & 78.5                                                          & 73.0                                                           & 93.0                                                             & 87.3                                                              & 93.4                                                                 \\
InfographicsVQA            & 4.6                  & 5.9                  & 40.8                                                              & 52.8                                                         & 39.3                                                          & 40.9                                                           & 67.7                                                             & 60.1                                                              & 73.9                                                                 \\
ChartQA                    & 1.4                  & 0.9                  & 59.0                                                              & 45.7                                                         & 41.7                                                          & 62.9                                                           & 64.1                                                             & 54.9                                                              & 75.4                                                                 \\
Visual7W                   & 4.0                  & 2.5                  & 47.7                                                              & 47.7                                                         & 49.5                                                          & 52.1                                                           & 61.6                                                             & 50.7                                                              & 55.4                                                                 \\
\rowcolor[RGB]{235, 242, 249} ScienceQA                  & 9.4                  & 5.2                  & 43.4                                                              & 41.1                                                         & 45.2                                                          & 52.1                                                           & 45.4                                                             & 46.7                                                              & 52.4                                                                 \\
\rowcolor[RGB]{235, 242, 249} VizWiz                     & 8.2                  & 1.7                  & 39.2                                                              & 49.8                                                         & 51.7                                                          & 44.6                                                           & 52.3                                                             & 41.9                                                              & 47.6                                                                 \\
\rowcolor[RGB]{235, 242, 249} GQA                        & 41.3                 & 43.5                 & 60.7                                                              & 53.0                                                         & 59.0                                                          & 61.2                                                           & 66.9                                                             & 69.0                                                              & 74.1                                                                 \\
\rowcolor[RGB]{235, 242, 249} TextVQA                    & 7.0                  & 4.6                  & 66.1                                                              & 78.9                                                         & 79.0                                                          & 73.5                                                           & 83.1                                                             & 73.0                                                              & 79.0                                                                 \\
\rowcolor[RGB]{235, 235, 235} All VQA                    & 9.1                  & 8.3                  & 54.9                                                              & 55.9                                                         & 57.4                                                          & 58.6                                                           & 62.9                                                             & 59.9                                                              & 67.8                                                                 \\ 
\midrule
\textbf{\textit{Retrieval (12 tasks)} }      &                      &                      &                                                                   &                                                              &                                                               &                                                                &                                                                  &                                                                   &                                                                      \\
VisDial                    & 30.7                 & 24.8                 & 73.3                                                              & 70.0                                                         & 83.0                                                          & 80.7                                                           & 80.5                                                             & 79.5                                                              & 82.7                                                                 \\
CIRR                       & 12.6                 & 39.1                 & 47.8                                                              & 43.5                                                         & 61.4                                                          & 54.0                                                           & 55.7                                                             & 54.3                                                              & 54.0                                                                 \\
VisualNews\_t2i            & 78.9                 & 50.7                 & 67.2                                                              & 70.6                                                         & 74.2                                                          & 73.3                                                           & 74.4                                                             & 73.7                                                              & 75.6                                                                 \\
VisualNews\_i2t            & 79.6                 & 21.1                 & 70.7                                                              & 74.1                                                         & 78.1                                                          & 76.9                                                           & 77.8                                                             & 73.7                                                              & 79.4                                                                 \\
MSCOCO\_t2i                & 59.5                 & 54.1                 & 70.6                                                              & 73.6                                                         & 78.6                                                          & 76.9                                                           & 76.4                                                             & 75.2                                                              & 74.6                                                                 \\
MSCOCO\_i2t                & 57.7                 & 40.0                 & 66.5                                                              & 71.1                                                         & 72.4                                                          & 73.7                                                           & 72.6                                                             & 71.2                                                              & 73.6                                                                 \\
NIGHTS                     & 60.4                 & 58.1                 & 66.1                                                              & 65.1                                                         & 68.3                                                          & 67.9                                                           & 67.4                                                             & 66.8                                                              & 65.3                                                                 \\
WebQA                      & 67.5                 & 43.0                 & 88.1                                                              & 85.0                                                         & 90.2                                                          & 89.6                                                           & 90.6                                                             & 86.9                                                              & 88.8                                                                 \\
\rowcolor[RGB]{235, 242, 249} FashionIQ                  & 11.4                 & 11.2                 & 12.9                                                              & 18.1                                                         & 54.9                                                          & 20.6                                                           & 22.2                                                             & 13.7                                                              & 22.2                                                                 \\
\rowcolor[RGB]{235, 242, 249} Wiki-SS-NQ                 & 55.0                 & 18.7                 & 56.6                                                              & 65.0                                                         & 24.9                                                          & 64.0                                                           & 73.3                                                             & 49.9                                                              & 65.7                                                                 \\
\rowcolor[RGB]{235, 242, 249} OVEN                       & 41.1                 & 1.6                  & 47.3                                                              & 65.9                                                         & 87.5                                                          & 58.2                                                           & 75.9                                                             & 67.4                                                              & 69.0                                                                 \\
\rowcolor[RGB]{235, 242, 249} EDIS                       & 81.0                 & 62.6                 & 79.9                                                              & 82.2                                                         & 65.6                                                          & 88.7                                                           & 80.8                                                             & 89.3                                                              & 91.3                                                                 \\
\rowcolor[RGB]{235, 235, 235} All Retrieval              & 53.0                 & 35.4                 & 62.3                                                              & 65.4                                                         & 69.9                                                          & 68.7                                                           & 70.6                                                             & 66.8                                                              & 70.2                                                                 \\ 
\midrule
\textbf{\textit{Visual Grounding (4 tasks)}} &                      &                      &                                                                   &                                                              &                                                               &                                                                &                                                                  &                                                                   &                                                                      \\
MSCOCO                     & 33.8                 & 22.1                 & 67.3                                                              & 64.6                                                         & 76.8                                                          & 75.2                                                           & 80.2                                                             & 81.1                                                              & 84.1                                                                 \\
\rowcolor[RGB]{235, 242, 249} RefCOCO                    & 56.9                 & 22.8                 & 84.7                                                              & 79.5                                                         & 89.8                                                          & 89.3                                                           & 92.1                                                             & 90.5                                                              & 94.6                                                                 \\
\rowcolor[RGB]{235, 242, 249} RefCOCO-Matching           & 61.3                 & 35.6                 & 79.2                                                              & 84.5                                                         & 90.6                                                          & 88.1                                                           & 92.8                                                             & 91.5                                                              & 92.8                                                                 \\
\rowcolor[RGB]{235, 242, 249} Visual7W-Pointing          & 55.1                 & 23.4                 & 86.8                                                              & 73.9                                                         & 77.0                                                          & 89.9                                                           & 89.5                                                             & 88.7                                                              & 88.8                                                                 \\
\rowcolor[RGB]{235, 235, 235} All Visual Grounding       & 51.8                 & 26.0                 & 79.5                                                              & 75.6                                                         & 83.6                                                          & 85.6                                                           & 88.7                                                             & 88.0                                                              & 90.1                                                                 \\ 
\midrule
\textit{\textbf{Final Score (36 tasks) }}    &                      &                      &                                                                   &                                                              &                                                               &                                                                &                                                                  &                                                                   &                                                                      \\
\rowcolor[RGB]{235, 235, 235} Avg. IND Scores             & 37.1                 & 31.0                 & 66.5                                                              & 65.8                                                         & 68.0                                                          & 70.5                                                           & 72.3                                                             & 70.6                                                              & 75.3                                                                 \\
\rowcolor[RGB]{235, 235, 235} Avg. OOD Scores             & 38.7                 & 23.7                 & 52.0                                                              & 60.1                                                         & 59.1                                                          & 57.9                                                           & 61.5                                                             & 58.4                                                              & 63.7                                                                 \\
\rowcolor[RGB]{235, 235, 235} Avg. Scores                 & 37.8                 & 27.8                 & 60.1                                                              & 63.3                                                         & 64.1                                                          & 64.9                                                           & 67.5                                                             & 65.2                                                              & 70.1                                                                 \\
\bottomrule
\end{tabular}
}
\end{table*}

\end{document}